\pdfoutput=1

\documentclass[11pt]{article}

\usepackage[final]{acl}

\usepackage{times}
\usepackage{latexsym}

\usepackage[T1]{fontenc}

\usepackage[utf8]{inputenc}

\usepackage{microtype}

\usepackage{inconsolata}

\usepackage{graphicx}
\usepackage{enumitem}
\usepackage{amsfonts}
\usepackage{amsmath}
\usepackage{multirow}
\usepackage{booktabs}
\usepackage[most]{tcolorbox}

\usepackage{xcolor}
\usepackage{xspace}
\usepackage{comment}

\newcommand{\ourapproach}{\textsc{Nemotron-CrossThink}\xspace}

\newcommand{\ours}{\ourapproach}

\newcommand{\rl}{\textsc{rl}\xspace}

\newcommand{\llm}{\textsc{llm}\xspace}

\newcommand{\nr}{\textsc{Natural Reasoning}\xspace}

\newcommand{\gpr}{\textsc{gpr}\xspace}

\newcommand{\mr}{\textsc{mr}\xspace}

\newcommand{\mmlu}{\textsc{mmlu}\xspace}
\newcommand{\amc}{\textsc{amc23}\xspace}
\newcommand{\orz}{\textsc{orz}\xspace}
\newcommand{\aime}{\textsc{aime24}\xspace}
\newcommand{\mcq}{\textsc{mcq}\xspace}
\newcommand{\open}{\textsc{open-ended}\xspace}
\newcommand{\gpqad}{\textsc{gpqa-diamond}\xspace}
\newcommand{\agieval}{\textsc{agieval}\xspace}
\newcommand{\supergpqa}{\textsc{supergpqa}\xspace}
\newcommand{\mmlupro}{\textsc{mmlu-pro}\xspace}

\newcommand{\cc}{CommonCrawl\xspace}

\newcommand{\qa}{\textsc{qa}\xspace}

\newcommand{\grpo}{\textsc{grpo}\xspace}

\newcommand{\mathall}{\textsc{math}\xspace}
\newcommand{\mathhard}{\textsc{math-500}\xspace}

\title{\ours: Scaling Self-Learning beyond Math Reasoning}

\author{Syeda Nahida Akter$^{2}$\thanks{Work done during internship at NVIDIA},~~ Shrimai Prabhumoye$^{1,3}$,~~ Matvei Novikov$^{1}$,~~ Seungju Han$^{1}$, \\\textbf{Ying Lin$^{1}$,~~ Evelina Bakhturina$^{1}$,~~ Eric Nyberg$^{2}$,~~ Yejin Choi$^{1}$,~~ Mostofa Patwary$^{1}$,}\\\textbf{Mohammad Shoeybi$^{1}$,~~ Bryan Catanzaro$^{1}$}\\
NVIDIA$^{1}$, Carnegie Mellon University$^{2}$, Boston University$^{3}$\\
\texttt{sakter@andrew.cmu.edu},~~\texttt{sprabhumoye@nvidia.com}}

\begin{document}


\maketitle

\begin{abstract}

Prior work has successfully applied Reinforcement Learning (\rl) to mathematical reasoning—where rules and correctness are well-defined. Yet, generalizing these methods to broader reasoning domains remains challenging due to limited data and the lack of verifiable rewards for unstructured domains
. In this work, we propose \ours, a framework that systematically incorporates multi-domain corpora
into \rl training to improve generalization across diverse reasoning tasks. \ours addresses key challenges by (1) 
combining data from varied sources
; (2) applying structured templates 
to control answer-space complexity; (3) filtering for verifiable answers; and (4) optimizing data blending strategies 
to utilize multi-source data effectively. 
This enables scalable and verifiable reward modeling beyond math 
and demonstrates improved accuracies on both math (\mathhard: +30.1\%, \amc: +27.5\%) and non-math reasoning benchmarks (\mmlupro: +12.8\%, \gpqad: +11.3\%, \agieval: +15.1\%, \supergpqa: +3.8\%). Moreover, \ours exhibits significantly improved response efficiency—
using 28\% fewer tokens for correct answers—highlighting more focused and effective reasoning. Through \ours, we demonstrate that integrating multi-domain, multi-format data in \rl leads to more accurate, efficient, and generalizable {\llm}s. All of our datasets are available on \href{https://huggingface.co/datasets/nvidia/Nemotron-CrossThink}{HuggingFace}.
\end{abstract}

\section{Introduction}\label{sec:introduction}

Large Language Models ({\llm}s) have demonstrated remarkable reasoning abilities across a wide range of tasks, with Reinforcement Learning (\rl) playing a key role in refining their deep thinking abilities \citep{OpenReasonerZero2025, aggarwal2025l1controllinglongreasoning, deepscaler2025, deepseekai2025deepseekr1incentivizingreasoningcapability, qin2024o1, huang2025o1replicationjourney, qwq32b2025}. Recent advances in \rl have been particularly successful in mathematical reasoning and coding, where well-defined rules and verifiable correctness enable effective reward modeling. Yet, extending these techniques to broader reasoning domains poses significant challenges, such as---limited training data for \rl due to the difficulty of defining verifiable rewards, and ensuring generalization across diverse tasks. 

\begin{figure}
    \centering  
    \includegraphics[width=\linewidth]{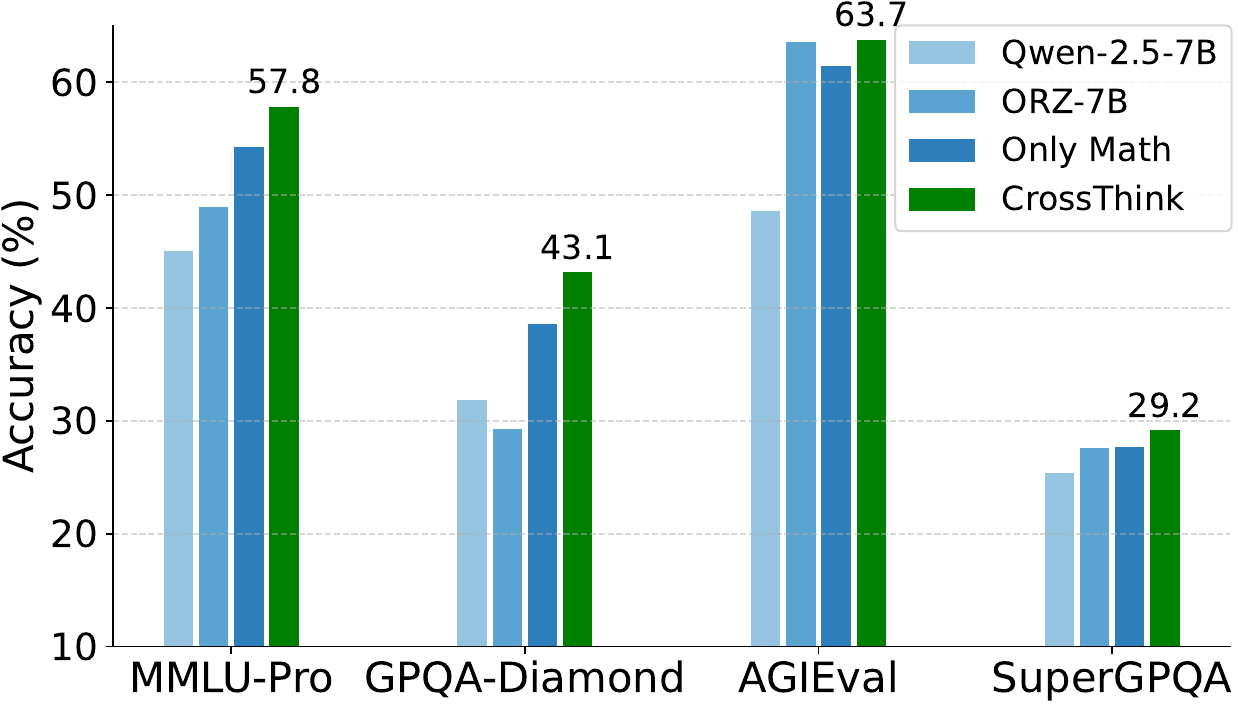}
    \caption{Employing self-learning with multi-domain data, \ours outperforms baseline models, including domain-specific training (Only Math) and Open-Reasoner-Zero (\orz-7B), achieving consistent gains across all reasoning tasks.}
  \label{fig:intro_img}
\end{figure}

Recent works \citep{OpenReasonerZero2025, deepscaler2025, cui2025process} have shown ways to diversify \rl training corpora by collecting datasets from multiple sources. However, they do not evaluate the relative importance of each source for reasoning 
or explore optimal data-blending strategies to maximize performance. 
Furthermore, prior research has largely focused on math reasoning, overlooking the role of non-math reasoning domains in \rl training for generalization in out-of-distribution domains. 
Reasoning process varies across domains and question types. For instance, math problem-solving follows a rule-based, structured, and symbolic approach \citep{dehaene2011number}, whereas reasoning in fields such as law, physics, social sciences, and history often relies on narrative structures, contextual knowledge, and heuristic strategies. Moreover, different question formats require distinct cognitive approaches---open-ended questions demand the generation of novel responses, while multiple-choice questions (\mcq) can often be solved by evaluating the given options and selecting the most appropriate answer. Incorporating a diverse range of reasoning domains and question types into \rl-based self-learning can enhance the broad reasoning ability of {\llm}s by exposing them to varied cognitive strategies and knowledge structures.

\begin{figure*}
  \centering
  \includegraphics[width=\textwidth]{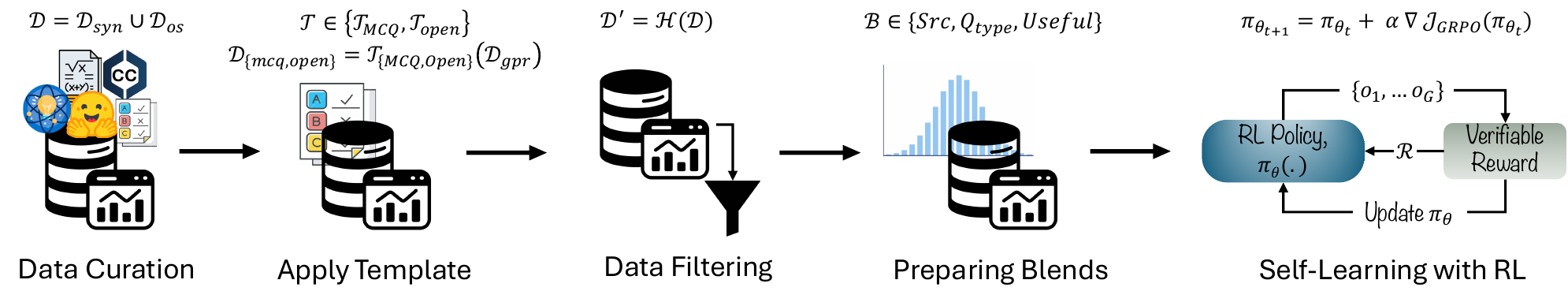}
  \caption{\textbf{\ours. } We (a) curate QA pairs from from CommonCrawl and open-source datasets, categorized into general-purpose reasoning ($\mathcal{D}_{gpr}$) and mathematical reasoning ($\mathcal{D}_{mr}$); (b) apply structured templates to convert data into \mcq and open-ended formats, promoting diverse reasoning trajectories; (c) filter out unverifiable or ill-formatted responses; (d) deploy \rl using Group Relative Policy Optimization (\grpo). The final reward is used to update the policy, iteratively improving the model’s reasoning capabilities across diverse domains.}
  \label{fig:cross_method}
\end{figure*}

In this work, we propose \ours, a systematic way to incorporate multi-domain corpora for \rl training that results in better generalization across a wide variety of tasks. As 
outlined in \autoref{fig:cross_method}, \ours comprises of phases that---(a) curate data from diverse sources, including synthetic data from web texts 
and open-source question-answer (QA) pairs on STEM, humanities, law, and social sciences (b) apply templates (\mcq/Open-Ended) to limit the answer-space 
(d) prepare blend
s to combine multi-domain data efficiently and (e) employ self-learning with \rl to refine reasoning capability in diverse domains.


We evaluate \ours along three axes: 
(1) the effectiveness of data blending strategies in self-learning 
(2) whether the blending impact amplifies by training with more complex data samples (3) the influence of question-answer templates on downstream accuracies. \ours demonstrates that the integration of multi-domain data with varied question formats in \rl 
boosts {\llm}'s reasoning 
across diverse 
domains (\autoref{fig:intro_img}). 
Notably, models trained with \ours not only achieve higher accuracy but also exhibit dynamic response strategies—generating concise answers for general-purpose questions and more detailed responses for math problems—thereby reducing inference cost while preserving task-specific rigor.
In addition, 
our approach addresses the challenge of designing scalable verifiable reward for non-deterministic domains by employing different templates on the curated data to limit the nuances in the answer space diversity. 
Furthermore, 
we explore a simple yet effective filtering approach to rank general purpose reasoning (\gpr) data based on complexity and show that training with harder samples further amplifies the impact of \rl across all domains.




In summary, our contributions are as follows:
\begin{itemize}[leftmargin=*]
\itemsep0em
\item We introduce {\ours}, a novel framework for incorporating multi-domain corpora into {\rl} training, enhancing generalization across diverse reasoning tasks with substantial gains on math (\mathhard: \textbf{+30.1\%}, \amc: \textbf{+27.5\%}) and non-math (\mmlupro: \textbf{+12.8\%}, \gpqad: \textbf{+11.3\%}, \agieval: \textbf{+15.1\%}, and \supergpqa: \textbf{+3.8\%}) benchmarks.
\item We demonstrate that applying question/answer templates to constrain output diversity leads to more stable reward modeling. Specifically, using a unified open-ended question format improves performance by \textbf{1.21\%} 
over mixed-format questions, while short-form answer templates outperform long-form ones by \textbf{1.20\%}.

 \item We show that math-only training is insufficient---blending multi-domain data in \rl  boosts average reasoning accuracy by \textbf{1.61\%} over math-only data and improves response efficiency by reducing token usage by \textbf{28\%}.
 \item We propose a simple yet effective model-driven filtering technique that selects harder samples by removing data solvable by smaller models. This leads to an additional \textbf{2.15\%} average accuracy gain for \texttt{Qwen-2.5-32B}, highlighting the scalability of our approach to larger models. 
 \item We release \textbf{287.4K} high-quality multi-domain data curated for verifiable reward modeling to support future research\footnote{\href{https://huggingface.co/datasets/nvidia/Nemotron-CrossThink}{https://huggingface.co/datasets/nvidia/Nemotron-CrossThink}}.
\end{itemize}

Applying \ours on different blends yields substantial improvement over base model (+8.55\%-13.36\% on average) across seven diverse \gpr and math benchmarks. The most effective blend—
2:1 ratio of \gpr to math data—achieves the highest average accuracy with a 13.36\% gain over baseline 
(\autoref{fig:intro_img}). 
Overall, these findings illustrate that thoughtful choices in data blending, scaling, formatting, and filtering are critical to the success of \rl with language models. 
We hope that \ours serves as a practical and extensible framework for leveraging multi-domain data to train more capable, reliable, and generalizable models under the \rl paradigm.




\section{\ours: Scaling Self-Learning Beyond Math}
\label{sec:method}


While mathematical reasoning benefits from clean, verifiable datasets, extending \rl to general-purpose reasoning is challenging due to the lack of structured, high-quality supervision.
To address this, 
we leverage web documents and open-source QA benchmarks to collect general-purpose reasoning (\gpr) data.
However, combining structured and unstructured domains introduces noise and ambiguity—particularly in open-ended formats—making it difficult to apply rule-based reward reliably. To mitigate this, we apply task-specific templates to unify formats, limiting answer space variability and enabling effective verifiable reward signals. Next, we apply a lightweight data filtering to discard unverifiable examples 
for stable and interpretable \rl training. Finally, we explore optimal data blending strategies 
to investigate how the inclusion of general-purpose reasoning data complements mathematical reasoning, ultimately leading to broader and more adaptive generalization in {\llm}s.


\paragraph{Data Curation. } As shown in \autoref{tab:data_dist}, we start with curating datasets from multiple sources to ensure diversity in the training data. Our training data $\mathcal{D}$ comprises of: 
\[
    \mathcal{D} = \mathcal{D}_{syn} \cup \mathcal{D}_{os}
\]
Here, $\mathcal{D}_{syn} \rightarrow$ synthetically generated from \cc (CC) \citep{gao2020pile800gbdatasetdiverse} and $\mathcal{D}_{os} \rightarrow$ open-source QA datasets. Each source of data further consists of QA pairs related to \gpr and math:
\begin{align*}
    \mathcal{D}_{\text{syn}} &\rightarrow \mathcal{D}_{\text{syn\_gpr}} \cup \mathcal{D}_{\text{syn\_mr}} \\
    \mathcal{D}_{\text{os}} &\rightarrow \mathcal{D}_{\text{os\_gpr}} \cup \mathcal{D}_{\text{os\_mr}}
\end{align*}   

\begin{itemize}[leftmargin=*]
    \item \textbf{General Purpose Reasoning}, $\mathcal{D}_{gpr}$: We collect open source QA datasets 
    ($\mathcal{D}_{os\_gpr}$)---Natural Reasoning \citep{yuan2025naturalreasoningreasoningwild28m} and \mmlu[Train] \citep{hendryckstest2021}  that span domains including STEM, Economics, Social Sciences, and more. To enhance diversity, we further synthesize QA pairs from CC documents 
    called \ours-\qa ($\mathcal{D}_{syn\_gpr}$).
    \[
    \mathcal{D}_{gpr}  \rightarrow \mathcal{D}_{syn\_gpr} \cup \mathcal{D}_{os\_gpr}
\]
    \item \textbf{Mathematical Reasoning}, $\mathcal{D}_{mr}$: 
    We combine open-source math datasets ($\mathcal{D}_{os\_mr}$): MATH \citep{hendrycksmath2021} and Numina-Math \citep{numina_math_7b}. We generate additional math problems 
    defined as \ours-\textsc{math} ($\mathcal{D}_{syn\_mr}$) to augment reasoning diversity.
    \[
    \mathcal{D}_{mr} \rightarrow \mathcal{D}_{syn\_mr} \cup \mathcal{D}_{os\_mr}\]
\end{itemize}


\paragraph{Applying Templates for Answer Space and Reasoning Diversity.}
General purpose reasoning benchmarks are often divided into two categories: (a) Multiple Choice Questions \citep{hendryckstest2021, wang2024mmlu} and (b) Open-Ended Questions \citep{zhong2023agieval}. Prior works overlooked these variations in the answer space for consistent reward design 
for tasks which are predominantly math \citep{OpenReasonerZero2025, aggarwal2025l1controllinglongreasoning, deepscaler2025}. We hypothesize that each question type elicits different thinking patterns, leading to diverse reasoning trajectories in the model. 
Therefore, 
we synthesize $\mathcal{D}_{gpr}$ using two templates: \( \mathcal{T}_{MCQ} \) - Multiple Choice Questions (\mcq), and \( \mathcal{T}_{Open} \) - Open-Ended questions. 
We convert the \mcq datasets (\mmlu) to open-ended by removing the options from the questions. 
\[
\mathcal{D}_{mcq} = \mathcal{T}_{MCQ}(\mathcal{D}_{gpr}), \quad \mathcal{D}_{open} = \mathcal{T}_{Open}(\mathcal{D}_{gpr})
\]


\begin{table}
\centering
\resizebox{\linewidth}{!}{%
\begin{tabular}{@{}lccc@{}}
\hline
\textbf{Data Source}      & \textbf{Category} & \textbf{Type}   & \textbf{Samples} \\
\hline
\mmlu[Train]        & \gpr      & \mcq             & 99,842 \\
\ours-\textsc{qa}      & \gpr        & \mcq             & 192,930 \\
\nr        & \gpr & \textsc{oe}      & 100,000 \\
NuminaMath               & \mr &   \textsc{oe}    & 87,350 \\
\ours-\textsc{math}           & \mr & \textsc{oe}      & 100,000 \\
Math                     & \mr & \textsc{oe}      & 8523 \\
\hline
\textbf{Total}           &               &  & \textbf{588,645} \\
\hline
\end{tabular}
}
\caption{Training data distribution by source and type. \textsc{oe}=Open-Ended; \textsc{gpr}=General-Purpose Reasoning; \textsc{mr}=Math Reasoning.} 
\label{tab:data_dist}
\vspace{-5mm}
\end{table}

Additionally, some \mcq questions are incomplete without options (e.g., \textit{Which of the following ways we can file taxes?}). We discard them to avoid confusion during answer generation. Finally, 
\[
\mathcal{D}_{gpr} = \mathcal{D}_{mcq}  \cup \mathcal{D}_{open}
\]








\paragraph{Data Filtering and Formatting.} To obtain high-quality data, we apply a series of filtering and formatting steps, $\mathcal{H}$, to remove samples that are infeasible to evaluate with 
rule-based reward. 
Specifically, for $\mathcal{D}_{mcq}$, we check whether the correct answer appears within the question text itself. Given a question-answer pair \((q, a^*)\) with answer choices \(\{a_1, a_2, \dots, a_n\}\), we discard a sample if 
\(a^* \notin \{a_1, a_2, \dots, a_n\}\). For $\mathcal{D}_{open}$, 
we discard samples that are challenging to evaluate with a rule-based reward function. Formally, we retain samples where $|w(a^*)| \leq 10$;
 \( w(a^*) \) represents the number of words in the answer \( a^* \). 

Lastly, for 
$\mathcal{D}_{mr}$, we remove entries that lack an associated answer, ensuring that all retained questions \( q \) have a valid response \( a^* \), i.e., we discard samples where $a^* = \emptyset$.
\[
\mathcal{D}' = \mathcal{H}(\mathcal{D}) = \left\{ (q, a^*, \{a_1, \dots, a_n\}) \in \mathcal{D} \right.
\]

\paragraph{Data Blending. } We study the impact of data diversity in three paradigms: 

\begin{itemize}[leftmargin=*,noitemsep]
    \item \textbf{Data Source: } 
    We observe the effect of data sources---$\mathcal{D}_{mr}$ and $\mathcal{D}_{gpr}$---by tuning their relative weights in the \rl training data.
    \item \textbf{Question Types: } We investigate the impact of question types in downstream tasks.
    \item \textbf{Data Usefulness: } 
    To analyze the contribution of each data source, we run \rl using individual data alone and then evaluate them across diverse downstream tasks. Based on their performances, we create a new blend. 
\end{itemize}
Based on these categories, we construct six blends, summarized in \autoref{tab:blend-description}, with their corresponding weight distributions detailed in \autoref{tab:blend_composition}.


\paragraph{Reinforcement Learning with \grpo. }
We begin with a pretrained large language model (\llm) $\mathcal{M}$ and a training blend $\mathcal{B}$, where each sample contains only the
input prompt and the final answer which is verifiable. We employ Group Relative Policy Optimization (\grpo) \citep{deepseek-math}. More details can be found in Appendix \ref{sec:grpo}.

\paragraph{Rule Based Reward Modeling. } 
To guide the \rl training, we employ a rule-based reward designed for verifiable evaluation. Similar to \citep{deepseekai2025deepseekr1incentivizingreasoningcapability}, we define the total reward function $\mathcal{R} = \mathcal{R}_{\text{acc}} \wedge \mathcal{R}_{\text{format}}$
as the combination of an accuracy reward \( \mathcal{R}_{\text{acc}} \) and a format reward \( \mathcal{R}_{\text{format}} \).
This implies that the output will get reward only when both the answer and the format are correct. Each reward is further detailed in Appendix \ref{sec:grpo}




\section{Data Synthesis}\label{sec:data_syn}

To obtain a balanced high-quality multi-domain data, we synthesize a high-quality question answering dataset spanning over math (\ours-\textsc{math}) and general purpose reasoning (\ours-\textsc{qa}) domain using publicly available datasets such as \cc \citep{gao2020pile800gbdatasetdiverse}. Additional details about the license can be found in Appendix \ref{sec:data_syn}. 

\subsection{\ours-\textsc{qa}}

We synthetically generate a large-scale multiple-choice question-answer (\mcq) dataset following two approaches below:

\subsubsection{SDG from Scratch}
\paragraph{Topic, Subtopic, and Difficulty Definition.}
We first define a broad set of topics, such as physics, biology, chemistry, and others. For each topic, we use \texttt{Nemotron-4-340B-Instruct}~\citep{nvidia2024nemotron4340btechnicalreport} to generate a list of popular subtopics. We also define multiple difficulty levels to ensure diversity and scale of the data.

\paragraph{Question Generation.}
We initially generate few-shot examples that demonstrate various levels of difficulty using \texttt{Nemotron-4-340B-Instruct}~\citep{nvidia2024nemotron4340btechnicalreport}. We later prompt \texttt{Qwen2.5} models~\citep{qwen2.5} along with the few-shot examples to generate a \mcq question based on the specified topic, subtopic, and difficulty. Each generated question is evaluated to ensure it follows the required format.

\paragraph{Augmentation.}
Similarly to the OpenMathInstruct~\citep{toshniwal2024openmathinstruct118millionmath} pipeline, we augment generated questions by prompting \texttt{Qwen2.5}~\citep{qwen2.5} models to create a question similar to or inspired by the original.

\paragraph{Benchmark Decontamination.} We perform decontamination against test sets of popular \mcq benchmarks such as \textsc{gpqa} \citep{rein2024gpqa}, \mmlu~\citep{hendryckstest2021}, and \mmlupro~\citep{wang2024mmlu}, following the approach suggested by \citet{yang2023rethinkingbenchmarkcontaminationlanguage}.

\paragraph{Solution Generation.}
We prompt DeepSeek-R1~\citep{deepseekai2025deepseekr1incentivizingreasoningcapability} to generate multiple reasoning traces per question. Since there are no ground-truth answers for the questions generated in earlier stages, we use majority voting over solutions to determine the most likely correct answer.

\subsubsection{SDG from Book}
We utilize \texttt{Qwen2.5-VL-72B-Instruct} \citep{qwen2.5-VL} for extracting text from textbooks (\textit{OpenStax}\footnote{\url{https://openstax.org/}} and \textit{An Introduction to Formal Logic}\footnote{\url{https://forallx.openlogicproject.org/forallxyyc.pdf}}), which was then manually checked for transcription accuracy. We employed \texttt{Mixtral-8x22B-Instruct-v0.1} \citep{jiang2024mixtralexperts} and \texttt{Qwen2.5-72B-Instruct} \citep{qwen2.5} to synthesize multiple-choice questions based on sections and key terms extracted from these textbooks. We prompted the models to generate questions with four distinct and plausible options, along with the correct answer and a justification for it. Subsequently, each generated question was evaluated using another model to ensure that every example is self-contained and accurate.

\subsection{\ours-\textsc{math}}


To construct \ours-\textsc{math}, we adopt an approach similar to \citet{ge2024scalingsyntheticdatacreation}. Specifically, we use web documents from Common Crawl 
and \texttt{Qwen2.5-72B-Instruct} \citep{qwen2.5} to generate personas. To promote diversity, we incorporate math skills introduced in \citet{didolkar2024metacognitive} and condition the \texttt{Qwen2.5-72B-Instruct} model on both the math skills and the personas. Finally, we use the \texttt{Qwen2.5-72B-Math-Instruct} \cite{yang2024qwen25mathtechnicalreportmathematical} model to generate the solutions. Prompt templates are shown in Figure~\ref{fig:persona_generation_1},~\ref{fig:persona_generation_2}, and~\ref{fig:persona_and_skill_to_problem}.

\section{Experimental Setup}\label{sec:exp_setup}

\paragraph{Training Details.} We adopt \texttt{Qwen2.5-7B} and \texttt{Qwen2.5-32B} \citep{qwen2.5} as 
$\mathcal{M}$, which demonstrate strong generalization capabilities across various reasoning tasks. We directly apply \grpo on $\mathcal{M}$ using the \texttt{veRL} framework\footnote{\url{https://github.com/volcengine/verl}}. 
We train $\mathcal{M}$ with key settings including a constant learning rate of 1e-6, a batch size and PPO mini batch size of 128 and a maximum  context length of 5000 tokens. Each generation step contains 128 unique prompts sampled from the dataset, and performing 8 rollouts with temperature and \texttt{top-p} both set to 1.0. We set KL coefficient to 0.001 in all experiments. We conduct training on 4 8 x NVIDIA-H100-80GB nodes, and each training takes approximately 48 GPUs hours. To further evaluate the efficacy of \ours across different model architectures and sizes, we conduct experiments with \texttt{Nemotron-H}, a 8B hybrid Mamba-Transformer model. The results of these experiments are presented in Appendix \ref{sec:general_model}. 

\paragraph{Evaluation Metrics.} We evaluate reasoning performance on diverse math and general-purpose benchmarks: 
\mathhard \citep{hendrycksmath2021}, \amc, test set of \mmlu \citep{hendryckstest2021}, \mmlupro \citep{wang2024mmlu}, \agieval \citep{zhong2023agieval}, \gpqad \citep{rein2024gpqa} and \supergpqa \citep{pteam2025supergpqascalingllmevaluation}. Notably, \supergpqa is a recent and rigorous benchmark designed to test the generalizability of {\llm}s across 285 graduate-level disciplines. 
Unlike existing benchmarks that concentrate on well-represented domains (e.g., math, law, physics), \supergpqa captures long-tail knowledge and includes a wide range of real-world professional disciplines, making it a reliable and discriminative frontier for evaluating generalizability in {\llm}s. 
We employ \texttt{vllm} \citep{kwon2023efficientmemorymanagementlarge} as the inference backend, with maximum response length of 5k. For the primary evaluation, we report accuracy averaged over three independent runs using greedy decoding. To further verify the robustness of \ours across different decoding strategies, we evaluate our best-performing models using sampling-based decoding in \autoref{tab:pass1_results}.

\section{Experiments and Results}\label{sec:experiments}

\begin{table*}[h!]
\centering
\resizebox{\textwidth}{!}{%
\begin{tabular}{lcccccccc}
\hline

\textbf{Data Source} & \textbf{\mmlu} & \textbf{\mmlupro} & \textbf{\gpqad} & \textbf{\agieval} & \textbf{\supergpqa} & \textbf{\mathhard} & \textbf{\amc} & \textbf{Avg} \\\hline
$\mathcal{M}$ & 74.20 & 45.00 & 31.82 & 48.59 &  25.36 & 48.30 &    40.00  &   44.75    \\
\hline
 \mmlu[Train]         & 69.76 & 38.50 & 32.83 & 47.66 & \textbf{27.69} & 22.00 & 5.00  & 34.78 \\
 \ours-\qa          & 70.45 & \textbf{52.41} & 30.81 & 52.10 & 24.57 & 54.20 & 35.00 & 45.65 \\
 Natural Reasoning  & 68.89 & 31.33 & 33.33 & 46.65 & 22.44 & 68.60 & 42.50 & 44.82 \\
 NuminaMath         & 72.94 & 52.05 & \textbf{33.84} & \textbf{54.39} & 26.97 & 76.20 & \textbf{55.00} & \textbf{53.06} \\
 \ours-\textsc{math}       & 53.99 & 28.08 & 18.69 & 45.69 & 16.92 & 77.20 & 50.00 & 41.51 \\
 Math               & 63.30 & 31.64 & 21.72 & 51.95 & 18.31 & \textbf{78.40} & 50.00 & 45.04 \\
\hline
\end{tabular}
}
\caption{\textbf{Results of Self-Learning on Individual Datasets.} Each row shows the downstream evaluation results after self-learning on a single data source. Results highlight the varying strengths of individual datasets across general-purpose and mathematical benchmarks.}
\label{tab:individual_results}
\end{table*}

\paragraph{Analyze the effect of Individual Datasets.} 
 To design an effective multi-source blend, we first assess the impact of each source on self-learning. This helps prioritize useful sources and downweight less effective ones. We employ 
 \rl using $\mathcal{M}$=\texttt{Qwen-2.5-7B} 
 on each dataset separately with a fixed training recipe for consistency. 
 Each model is trained for 250 steps and evaluated on the final checkpoint.

As shown in \autoref{tab:individual_results}, different datasets have varying impacts on downstream accuracies across reasoning benchmarks. 
\texttt{NuminaMath} yields the highest overall average, outperforming the baseline ($\mathcal{M}$) by over 8.30\%. While particularly strong on math tasks like \mathhard and \amc, it also generalizes well to broader reasoning benchmarks.
\ours-\qa demonstrates a $\sim$1.0\% improvement over baseline with stronger accuracy in \mmlupro, \agieval and \mathhard tasks, suggesting that synthetically generated instruction-style data can generalize well when aligned with benchmark distributions. 
\texttt{Natural Reasoning}, despite modest scores on language-rich benchmarks, delivers a strong average, driven by high scores in \mathhard and \amc. 
This indicates that reasoning-focused datasets, even if less formatted, can contribute meaningfully in math-adjacent tasks. 
In contrast, \ours-\textsc{math} performs well on math but generalizes poorly to other domains.
Finally, \mmlu[Train] underperforms across most tasks, specifically in math domains, suggesting that self-learning with raw \mmlu[Train] data alone is insufficient for generalization.
However, it excels on \supergpqa, which spans cross-disciplinary reasoning, highlighting its potential in capturing broad conceptual knowledge and supporting transfer to long-tail domains---making it a valuable component when targeting general-purpose reasoning benchmarks.
While preparing $\mathcal{B}_{score}$, we weight datasets based on their average accuracy—prioritizing sources like \ours-\qa and \texttt{NuminaMath}, while downweighting less effective ones like \mmlu[Train].


\paragraph{Analysis across Blends.} 
To show the distinction between natural distribution and selective weighting of domains, we prepare $\mathcal{B}_{nd}$, which samples data in proportion to each dataset's original size. 
Next, to analyze the impact of within-domain vs. cross-domain training, we introduce a \texttt{Single Source} category with two domain-specific blends:  
$\mathcal{B}_{only\_mr}$ and $\mathcal{B}_{only\_gpr}$, using only $\mathcal{D}_{mr}$ and $\mathcal{D}_{gpr}$ respectively. 
We further compare our approach 
with a recent math-centric self-learning approach, \textsc{Open-Reasoner-Zero} (\orz) \citep{OpenReasonerZero2025}---which achieved 
strong math accuracy using combination of math data. 
For fair comparison, we evaluate \orz-7B using our eval setup.

\begin{table*}[htbp]
\centering
\resizebox{\textwidth}{!}{%
\begin{tabular}{lcccccccccc}
\hline
 \textbf{Model} & \textbf{Category}& \textbf{Blend} & \textbf{\mmlu} & \textbf{\mmlupro} & \textbf{\gpqad} & \textbf{\agieval} & \textbf{\supergpqa} & \textbf{\mathhard} & \textbf{\amc} & \textbf{Avg} \\
\hline
$\mathcal{M}$  & & & 74.20 & 45.00 & 31.82 & 48.59 & 25.36 & 48.30 & 40.00 & 44.75\\
\orz & & & 73.20 &	48.90	&29.30	&63.49	&27.60	&81.40	&62.50& 55.20\\\midrule
\multirow{7}{*}{\rotatebox[origin=c]{90}{\textsc{CrossThink}$^{*}$}} & & $\mathcal{B}_{nd}$ & 73.18 & 54.81 & 38.07 & 59.99 & 26.54 & 77.00 & 60.00 & 55.66\\\cmidrule{2-11}
& \multirow{2}{*}{Data Source} & $\mathcal{B}_{mr\uparrow}$ &  74.85 & 55.51 &  40.10 & 61.47 & 26.81 & 77.80 &  67.50 & 57.72\\
& & $\mathcal{B}_{gpr\uparrow}$ &  \textbf{74.94}	& \textbf{57.82}&	38.58	& \textbf{63.71}	& \textbf{29.16} &  77.60	&65.00   &  \textbf{58.12} \\\cmidrule{2-11}
& \multirow{2}{*}{Question Types} & $\mathcal{B}_{mcq\uparrow}$ & 74.26 & 55.77 & 39.59 & 62.54 &  28.05 & 78.00 & 60.00 & 56.89 \\
& & $\mathcal{B}_{open\uparrow}$ & 74.46 & 55.82 &  \textbf{43.15} & 61.28 & 26.82 & 78.40 & 62.50 & 57.49\\\cmidrule{2-11}
& Data Usefulness  & $\mathcal{B}_{score}$ & 74.70 &  56.16 & 40.10 & 59.80 & 27.37 & 78.00 & 62.50 & 56.95\\\midrule 
& \multirow{2}{*}{Single Source} & $\mathcal{B}_{only\_mr}$ & 74.24 & 54.26 & 38.58 & 61.39 & 27.69 &  \textbf{78.60} &  \textbf{70.00} & 57.82\\
& & $\mathcal{B}_{only\_gpr}$ & 72.77 & 52.06 & 37.06 & 56.56 & 27.44 & 72.20 & 55.00 & 53.30\\
\hline
\end{tabular}
}%
\caption{\textbf{Results of \ours-7B across Blends.} 
$\mathcal{B}_{gpr\uparrow}$ achieves the highest overall average accuracy, outperforming domain-specific and naturally sampled blends---underscoring the benefit of self-learning with diverse reasoning data. (*) Due to the space shortage, we use \textsc{*CrossThink} to refer \ours.}
\label{tab:7b_blends}
\end{table*}

As shown in \autoref{tab:7b_blends}, each blend outperforms $\mathcal{M}$ by a significant margin. 
$\mathcal{B}_{nd}$ yields a 13\%  average improvement over $\mathcal{M}$, suggesting that simple
data diversity---even without rebalancing---can be beneficial. 
$\mathcal{B}_{gpr\uparrow}$ 
achieves the highest overall average, with the strongest results across most benchmarks (e.g., \mmlupro: +12.82\%, \agieval: +15.12\%). 
Notably, it outperforms \orz by $\sim$5\% on average.
While $\mathcal{B}_{only\_mr}$ performs slightly better on math, 
it lags $\sim$3–4\% behind $\mathcal{B}_{gpr\uparrow}$ on non-math reasoning tasks 
such as \agieval, \supergpqa, and \mmlupro. 
The trend also holds for \orz. 
Our analysis with sub-category accuracies in \autoref{sec:subcat_abl} 
reveals that $\mathcal{B}_{gpr\uparrow}$ shows large relative gains in non-math categories while gains in math subcategories are either negligible or even favor $\mathcal{B}_{gpr\uparrow}$ in some tasks. This highlights that multi-domain data offers strong cross-domain transfer with minimal compromise on math accuracy, making it more versatile.


Both $\mathcal{B}_{mcq\uparrow}$ and $\mathcal{B}_{open\uparrow}$ 
show consistent gains, with the latter achieving a slight edge (+0.6\% on average) with stronger results on math tasks.
Since math problems are inherently open-ended in structure, 
highlighting more open-ended domains aligns with the format and reasoning demands of math tasks---
leading to better generalize to both general purpose reasoning (\gpr) and math tasks.
Despite outperforming $\mathcal{M}$, $\mathcal{B}_{score}$ is overall worse than $\mathcal{B}_{mr\uparrow}$ or $\mathcal{B}_{only\_mr}$. This gap arises because $\mathcal{B}_{score}$ assigns weights based on average scores, without accounting for task-specific strengths. 
For example, \texttt{Math} and \ours-\textsc{math} are overrepresented due to math performance, while datasets like \texttt{MMLU} or \texttt{Natural Reasoning}, which excel in general reasoning, are underweighted. In contrast, domain-aware blends selectively prioritize datasets based on their utility within specific domains, leading to more effective coverage and stronger scores across both math and \gpr tasks.  


In \textit{Single Source} vs. multi-domain analysis, $\mathcal{B}_{only\_mr}$ achieves the highest average math score,
 ranking as the second-best blend overall in terms of average accuracy. 
In contrast, while $\mathcal{B}_{only\_gpr}$ outperforms $\mathcal{M}$, it underperforms in math tasks and trails 4.2\% on average across non-math reasoning tasks, 
despite being tailored for \gpr.
This counterintuitive finding suggests that to obtain maximum gain in \gpr tasks we need to include math problems in the training. 
As discussed earlier, $\mathcal{B}_{gpr\uparrow}$ gets the best average reasoning accuracy which consists of both math and \gpr domains. 
This confirms that math data alone is transferable to structured reasoning tasks, whereas \gpr data is less effective when isolated.

\section{Ablations}\label{sec:ablations}

\paragraph{\ours is token efficient in responses.}

\begin{figure}
  \centering
  \includegraphics[width=\linewidth]{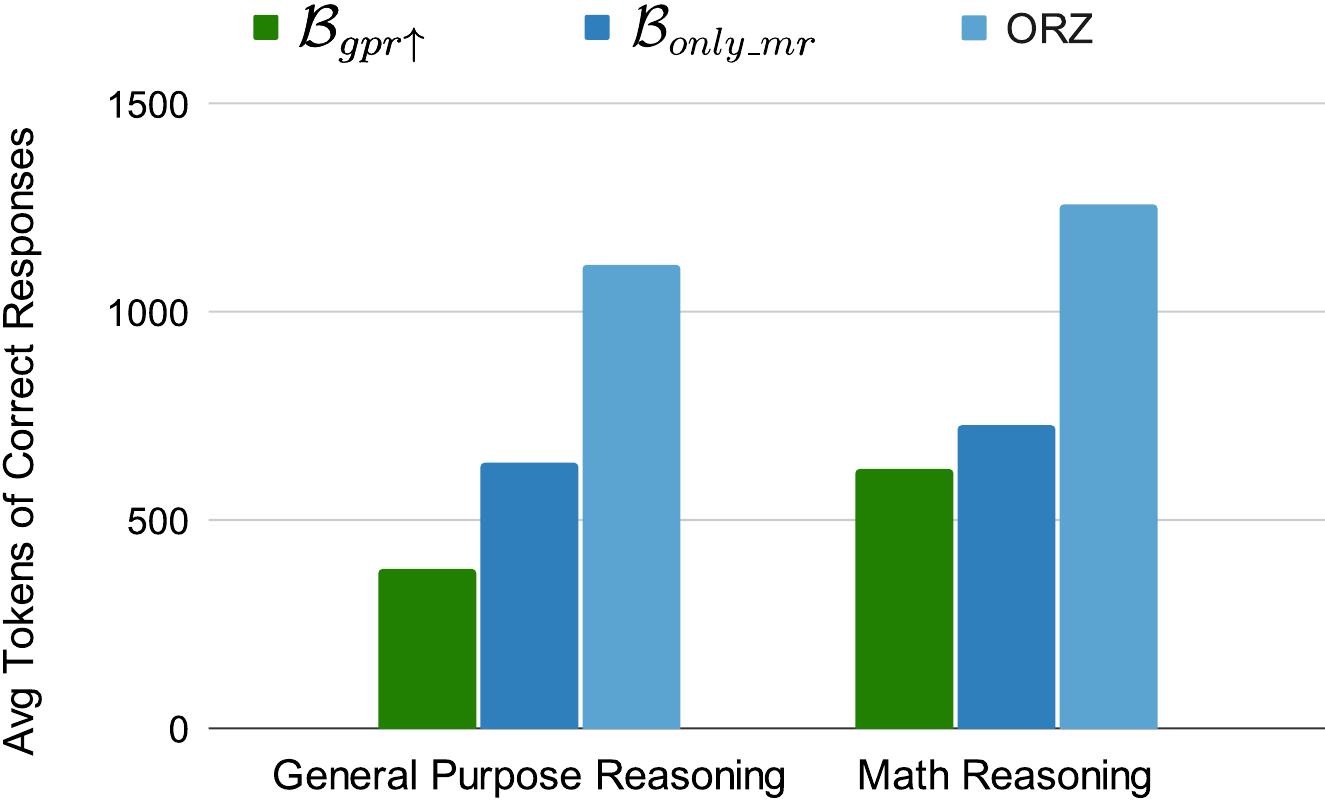}
  \caption{Token efficiency comparison of models trained on $\mathcal{B}_{gpr\uparrow}$ (multi-domain blend) and two single domain blends ($\mathcal{B}_{only\_mr}$ and \orz).}
  \label{fig:token_efficiency}
\end{figure}

To further understand the influence of multi-domain data in response generation, we compare the average token lengths of correct and incorrect responses between models trained on two blends: $\mathcal{B}_{gpr\uparrow}$ and $\mathcal{B}_{only\_mr}$. 
As shown in \autoref{fig:token_efficiency}, on general-purpose reasoning (\gpr) benchmarks, $\mathcal{B}_{gpr\uparrow}$ consistently outperforms $\mathcal{B}_{only\_mr}$ and \orz \citep{OpenReasonerZero2025}, not only in accuracy (as shown in \autoref{tab:7b_blends}) but also in response efficiency—producing correct answers with significantly fewer tokens\footnote{Detailed categorization per task is shown in Appendix \ref{sec:token-efficiency}.}.
For instance, on \mmlu, the average token count for correct responses is 229 for $\mathcal{B}_{gpr\uparrow}$, compared to 351 for $\mathcal{B}_{only\_mr}$. 
This demonstrates that exposure to multi-domain data enables the model to internalize a more efficient reasoning strategy, leading to both improved performance and reduced inference cost.

In contrast, on math-specific benchmarks, $\mathcal{B}_{only\_mr}$ and \orz perform slightly better in accuracy, as expected due to domain alignment. Interestingly, correct responses are generally longer than \gpr tasks as solving math problems inherently requires detailed, multi-step derivations, hypothesis exploration, verification and refinement. 
Despite this, the $\mathcal{B}_{gpr\uparrow}$ shows its adaptability by generating longer responses for math tasks and shorter ones for \gpr tasks—indicating a dynamic response strategy learned through multi-domain training. 
As shown in \autoref{tab:token_stats}, $\mathcal{B}_{gpr\uparrow}$ 
increases its average tokens by 62\% when generating responses for math tasks (Mean Tokens=622) as opposed to \gpr tasks (Mean Tokens=385). 
Whereas, $\mathcal{B}_{only\_mr}$ increases 
by 14\% (Mean Tokens=731 for math and Mean Tokens=639 for \gpr tasks) showing a much smaller dynamic range. This trend is also mirrored in \orz 
which shows an even smaller increase (12\%) in average token length across domains.

This adaptive behavior highlights a key strength of multi-domain training: it equips the model with the flexibility to tailor its response style to the nature of the task. 
By learning from a diverse range of domains, $\mathcal{B}_{gpr\uparrow}$ learns to reason efficiently---across all tasks, $\mathcal{B}_{gpr\uparrow}$ uses on average 28\% fewer tokens for correct responses than $\mathcal{B}_{only\_mr}$---producing compact yet accurate answers where appropriate, and detailed ones when necessary. Our experiment with a new model family \texttt{Nemotron-H} also reveal the similar trend (\autoref{tab:nemotron_tokens})---confirming that the efficiency in thinking traces achieved through \ours is architecture and model-agnostic.





\paragraph{Data Format Study: Question and Answer Templates.} 
To 
examine the training data formatting 
effect on model performance, we conduct two controlled studies focused on question and answer template design. 
In \autoref{tab:7b_blends}, we observe that $\mathcal{B}_{open\uparrow}$ outperforms $\mathcal{B}_{mcq\uparrow}$, suggesting that models trained on more open-ended data generalize better across benchmarks. 
This motivated us to investigate whether converting all questions into a unified open-ended format leads to better performance.

In \textbf{\textit{Question Template Study}}, 
we use the natural distribution blend ($\mathcal{B}_{nd}$) and only perturb the question template. 
To generate the open-ended variant, we remove the answer options from {\mcq}s, prompting the model to produce an answer without selecting from predefined choices. 



\begin{table}[htbp]
\centering
\resizebox{\linewidth}{!}{%
\begin{tabular}{lccc}
\toprule
\textbf{Question Type} & \textbf{GPR Avg} & \textbf{Math Avg} & \textbf{Total Avg} \\
\midrule
\mcq+\open & 50.52	& 68.50	& 55.66\\
\open & \textbf{51.30} & \textbf{70.80} & \textbf{56.87}\\
\bottomrule
\end{tabular}
}%
\caption{\textbf{Impact of Question Format.} Converting all questions to open-ended format improves accuracy across benchmarks, reducing reliance on option guessing and encouraging deeper reasoning (Appendix \ref{sec:breakdown_perf}).}
\label{tab:q_template}
\end{table}

From \autoref{tab:q_template}, the open-ended setting surpasses the mixed-format one on nearly all tasks, achieving 1.21\% higher average score. It yields notable gains on reasoning-intensive and \mcq-formatted benchmarks such as \mmlu, \supergpqa, and \gpqad. This result may be attributed to the inherent structure of \mcq questions, where random guessing can yield an accuracy of approximately 25\% in \mmlu and \gpqad where we have four options. In contrast, open-ended questions eliminate this guessing advantage, compelling the model to rely heavily on reasoning to arrive at a correct answer. By reducing the likelihood of reward hacking through random option selection, the open-ended format encourages more robust reasoning and leads to improved generalization.

In \textbf{\textit{Answer Template Study}}, we investigate how the format of \mcq-style output labels influences training. 
We compare two answer templates: \textit{Long} - the model is trained to generate both the option label and its corresponding description (e.g., \texttt{(A) Sky is blue}), and \textit{Short} - the model is trained to output only the option label (e.g., \texttt{A}). Here, we use the $\mathcal{B}_{only\_gpr}$ blend, which primarily consists of \mcq datasets (\autoref{tab:data_dist}), making it ideal for analyzing the effects of answer formatting in this setting.



\begin{table}[htbp]
\centering
\resizebox{\linewidth}{!}{%
\begin{tabular}{lccc}
\toprule
\textbf{Answer Type} & \textbf{GPR Avg} & \textbf{Math Avg} & \textbf{Total Avg}\\
\midrule
Long & 49.18	& \textbf{63.60}	& 53.30\\
Short & \textbf{50.95}	& 63.35	& \textbf{54.50}\\
\bottomrule
\end{tabular}
}
\caption{\textbf{Impact of Answer Format.} Using short-form answers improves accuracy by reducing output ambiguity and avoiding penalization from rigid reward functions in rule-based training (Appendix \ref{sec:breakdown_perf}).}

\label{tab:a_template}
\end{table}

As shown in \autoref{tab:a_template}, the short answer template outperforms the long-form variant, with a 1.20\% gain in average accuracy. The trend holds for both \gpr and math benchmarks. These results suggest that reducing the complexity of the output space helps minimize ambiguity and allows the model to better align its predictions with the structure of the question. Furthermore, when training with long-form answers using a rule-based reward (e.g., exact string matching), the model is often penalized for minor deviations in phrasing, even when the correct option is selected. 
This introduces noisy supervision and may hinder learning. While this issue could be mitigated by designing a more flexible reward function (e.g., \llm-as-a-Judge), 
we aim to keep our approach simple and interpretable. As such, we adopt a naive rule-based reward for clarity and reproducibility, and leave more sophisticated reward designs for future investigation.


\paragraph{Difficulty Filtering.} 
High-quality data is a key factor in self-learning to ensure efficient and stable learning. 
Recent works \citep{OpenReasonerZero2025, deepscaler2025, cui2025process, zeng2025simplerlzooinvestigatingtamingzero, fatemi2025concisereasoningreinforcementlearning} 
investigate data selection based on question complexity, showing that 
training on harder questions improves downstream accuracy. However, their approach relies on datasets 
with predefined difficulty scores. In this work, we explore a simple approach to estimate question difficulty for \gpr datasets that do not come with explicit difficulty labels. 
Specifically, we label questions as `difficult' if they are answered incorrectly by a smaller model (\texttt{Qwen-2.5-7B}) in a zero-shot setting and filter out the `easy' questions. The intuition is that questions easily answered by a base model are likely to be knowledge-based or shallow in reasoning depth, whereas those it fails on are likely to require deeper reasoning or broader generalization. 
We construct two versions of our training dataset $\mathcal{B}_{gpr\uparrow}$---an unfiltered set containing all questions, and a filtered set ($\mathcal{B}_{f(gpr)\uparrow}$) that retains only the difficult samples---and use them to train separate instances of a larger $\mathcal{M}$ = \texttt{Qwen-2.5-32B}. 

\begin{table}[htbp]
\centering
\resizebox{\linewidth}{!}{%
\begin{tabular}{lcccc}
\toprule
\textbf{Model} & \textbf{Blend} & \textbf{GPR Avg} & \textbf{Math Avg} & \textbf{Total Avg}\\
\midrule
Qwen-2.5-32B & &54.95& 52.78  &54.33\\\midrule
\multirow{2}{*}{\ours-32B} &  $\mathcal{B}_{gpr\uparrow}$ & 62.20	&74.95	&65.84\\
&  $\mathcal{B}_{f(gpr)\uparrow}$ & \textbf{63.39}	& \textbf{79.50} &	\textbf{67.99}\\
\bottomrule
\end{tabular}
}
\caption{\textbf{Difficulty-Based Filtering.} Filtering $\mathcal{B}_{gpr\uparrow}$ to retain only hard examples (\smash{$\mathcal{B}_{f(gpr)\uparrow}$}) yields consistent gains across all tasks, highlighting the effectiveness of selective training on challenging data (Appendix \ref{sec:breakdown_perf}).}

\label{tab:data_filter}
\end{table}

According to \autoref{tab:data_filter}, 
this filtering approach results in consistent performance gains across all evaluated benchmarks. While both filtered and unfiltered models outperform $\mathcal{M}$,  
$\mathcal{B}_{f(gpr)\uparrow}$ achieves the highest accuracy on every task. The gains are especially prominent in complex benchmarks such as \mmlupro, \gpqad, \agieval, and \amc, where $\mathcal{B}_{f(gpr)\uparrow}$ improves by up to 2–8\% over $\mathcal{B}_{gpr\uparrow}$. 
On average, filtering boosts overall accuracy by 2.15\%, a notable gain considering that it comes from training on fewer but harder examples. This suggests that selectively training on challenging examples can yield more robust and generalizable models, likely due to stronger gradient signals and a focus on harder-to-learn reasoning patterns. 


\section{Related Work}
\label{sec:related_works}

\paragraph{Reasoning in \llm.}
Large Language Models have achieved strong performance on various NLP tasks, with Chain-of-Thought (CoT) prompting \citep{wei2022chain} enabling multi-step reasoning across domains like math, science, and programming. Long CoT \citep{openai2024gpt4o} further enhances reasoning by introducing behaviors such as reflection, verification, and correction, with strong scaling properties. Models like QwQ \citep{qwq2024, qwq32b2025}, DeepSeek-R1 \citep{deepseekai2025deepseekr1incentivizingreasoningcapability}, Kimi k1.5 \citep{kimiteam2025kimik15scalingreinforcement}, and InternThinker \citep{cai2024internlm2technicalreport} leverage Long CoT with RL to boost reasoning performance. Smaller models like Open-Reasoner-Zero \citep{OpenReasonerZero2025}, Open-R1 \citep{openr1}, O1 \citep{qin2024o1, huang2025o1replicationjourney}, s1 \citep{muennighoff2025s1simpletesttimescaling}, and LIMO \citep{ye2025limoreasoning} also benefit from Long CoT via distillation.

\paragraph{Data Sampling in \rl.}
Recent work explores mixing data from multiple sources in \rl to improve reasoning diversity and generalization in \rl \citep{OpenReasonerZero2025, deepscaler2025, zeng2025simplerlzooinvestigatingtamingzero, wen2025lightr1curriculumsftdpo}. However, they primarily focus on math due to the ease of designing verifiable rewards. Sampling strategies often rely on question complexity or algorithmic verifiability. 
\citet{xie2025logicrlunleashingllmreasoning}, uses synthetic puzzles to control difficulty. However, these methods remain limited to structured domains like math. \citet{yeotong2025longcot} reports the best \mmlupro scores by blending multi-domain \cite{yue2024mammoth2} data, though majority of it is math-focused, leaving unclear the contribution of non-math data. 
Despite these efforts, the impact of including non-math domains—like law, social science, or commonsense reasoning—remains underexplored. \ours is the first systematic framework to incorporate multi-domain, multi-format data into \rl, introducing verifiable rewards for non-deterministic tasks and demonstrating that diverse blends lead to {\llm}s that reason more broadly, adapt dynamically, and think more efficiently.

\section{Conclusion}

We present \ours, a simple and scalable framework for improving the generalization abilities of {\llm}s through \rl with multi-domain corpora. By combining multi-domain data, structured templates, and difficulty-aware filtering, \ours enables consistent gains across both general-purpose (+3.8–15.1\%) and mathematical (+27.5–30.1\%) benchmarks---using 28\% fewer tokens for correct responses.
Importantly, these benefits persist across model scales and task types, demonstrating that data diversity, not just data volume, is key to broader reasoning capabilities. \ours offers a practical recipe for building more generalizable, efficient, and reliable {\llm}s under the \rl paradigm—paving the way for scalable self-learning beyond math.

\section{Limitations}
\label{sec:limitations}

While \ours demonstrates strong improvements in reasoning accuracy, adaptability, and efficiency, there is still room for improvement. As discussed in Section \ref{sec:ablations}, the reward modeling framework used in this work is rule-based and relatively simplistic. Specifically, it relies on exact string matching for correctness and formatting verification, which can be brittle for open-ended responses. For example, if the ground truth is \texttt{(A) Sky is blue}, and the model predicts \texttt{(A) the sky is generally blue most times}, the answer is semantically correct but still receives a negative reward. This limitation improperly affects general-purpose reasoning tasks with inherently more diverse and less deterministic answer spaces. Future work could incorporate more flexible, semantics-aware reward functions, such as fuzzy matching, entailment scoring, embedding-based similarity metrics, or \texttt{\llm-as-a-Judge}, to better align reward signals with human judgment. Additionally, we did not perform extensive hyperparameter tuning for \rl training. All models were trained using fixed schedules and standard values for learning rate, KL coefficients, and rollout configurations. Moreover, scaling \rl training for longer steps and dataset are computationally expensive which constrained us to deploy all runs for a fixed number of steps (650 steps). Recent works \citep{yu2025dapoopensourcellmreinforcement, aggarwal2025l1controllinglongreasoning, deepscaler2025} shows improvement over naive \grpo by adjusting clip ratio, dynamic sampling, number of rollouts, context length. While our results are strong under fixed conditions, additional gains may be possible with better-tuned training regimes by exploiting hyperparameters.

\section{Ethical Considerations}

Reinforcement learning strongly incentivizes the reasoning capabilities of \llm{s}, enabling models to perform better across complex tasks. However, this process can also inadvertently amplify existing biases present in the base model or introduced through reward modeling and data selection. In this work, we primarily focus on scaling and diversifying reasoning ability across domains and do not explicitly address fairness, bias mitigation, or value alignment.  Future work should systematically evaluate how reinforcement learning affects the model’s behavior across sensitive axes such as gender, race, and geopolitical context—especially in open-ended, non-verifiable tasks.

\bibliography{custom}

\appendix
\section{Reinforcement Learning}\label{sec:grpo}
We utilize Group Relative Policy Optimization (\grpo) \citep{deepseek-math} as our \rl algorithm. Unlike \textsc{PPO} \citep{schulman2017proximalpolicyoptimizationalgorithms}, \grpo does not use a separate critic model and instead estimates the baseline from group scores, improving efficiency and reducing memory.  For each question $q$, \grpo samples a group of outputs $o_1, o_2, ..., o_G$ from the old policy $\pi_{\theta_{old}}$ and then
optimizes the policy model $\pi_\theta$ by maximizing the following objective:

\begin{align*}
    x_{i,t} = \frac{\pi_{\theta}(o_{i,t} | q, o_{i, <t})}{\pi_{\theta_{\text{old}}}(o_{i,t} | q, o_{i, <t})}
\end{align*}
    
\begin{small}
\begin{align*}
\mathcal{J}_{\text{\grpo}}(\theta) = 
\; \mathbb{E}\left[q \sim P(Q), \{o_i\}_{i=1}^{G} \sim \pi_{\theta_{\text{old}}}(O | q)\right] \\
\times \frac{1}{G} \sum_{i=1}^{G} 
\frac{1}{|o_i|} \sum_{t=1}^{|o_i|}
\Bigg[ \min \Bigg(x_{i,t} 
 \hat{A}_{i,t},\\
\text{clip} \Bigg(x_{i,t}, 1 - \epsilon, 1 + \epsilon \Bigg) \hat{A}_{i,t}
\Bigg) \\
\quad - \beta D_{\text{KL}} \big( \pi_{\theta} \| \pi_{\text{ref}} \big)
\Bigg]
\end{align*}
\end{small}

\begin{align*}
    D_{\text{KL}} \left[ \pi_{\theta} \| \pi_{\text{ref}} \right] = 
    \frac{\pi_{\text{ref}}(o_{i,t} | q, o_{i,<t})}{\pi_{\theta}(o_{i,t} | q, o_{i,<t})}\\
    - \log \frac{\pi_{\text{ref}}(o_{i,t} | q, o_{i,<t})}{\pi_{\theta}(o_{i,t} | q, o_{i,<t})} - 1.
\end{align*}

where $\epsilon$ and $\beta$ are hyperparameters, and $\hat{A}_{i,t}$
is the advantage, computed using a group of rewards $\{r_1, r_2, ..., r_G\}$
corresponding to the outputs within each group:
\[
 \hat{A}_{i,t} = \frac{r_i - \text{mean}(\{r_1, r_2, ..., r_G\})}{\text{std}(\{r_1, r_2, ..., r_G\})}
\]

\paragraph{Defining Rewards.} We combine accuracy reward ($\mathcal{R}_{\text{acc}}$) and format reward ($\mathcal{R}_{\text{format}}$) to estimate the final reward:

\textbf{Accuracy Reward:} The accuracy reward evaluates correctness based on whether the model's response \( p \) is similar to the ground truth solution \( a \) to satisfy the correctness criteria:
\[
    \mathcal{R}_{\text{acc}}(p, a) =
    \begin{cases}
        1, & \text{if equal}(p, a), \\
        0, & \text{otherwise}.
    \end{cases}
\]


\textbf{Format Reward:} The format reward ensures the response \(a\) is structured according to predefined tags, where the reasoning will reside in `<think></think>' tokens and the final answer will be shown inside \texttt{\textbackslash boxed\{\}}:
\[
    R_{\text{format}}(a) =
    \begin{cases}
        1, & \text{if } F(a), \\
        0, & \text{otherwise}.
    \end{cases}
\]

where \( F(a) \) returns \texttt{True} if \( a \) is correctly formatted and \texttt{False} otherwise.

\section{Additional Details on Data Synthesis}\label{sec:data_syn}

Our dataset is intended to be used by the community to deploy reinforcement learning with {\llm}s which is licensed under the Creative Commons Attribution 4.0 International License (CC BY 4.0)\footnote{\url{https://creativecommons.org/licenses/by/4.0/legalcode}}. The data may be used to train and evaluate. This dataset contains synthetic data created using \texttt{Qwen/Qwen2.5-Math-72B}, \texttt{Qwen2.5-72B-Instruct}. If this dataset is used to create, train, fine tune, or otherwise improve an AI model, which is distributed or made available, such AI model may be subject to redistribution and use requirements in the Qwen License Agreement\footnote{\url{https://huggingface.co/Qwen/Qwen2.5-Math-72B/blob/main/LICENSE} and \url{https://huggingface.co/Qwen/Qwen2.5-72B-Instruct/blob/main/LICENSE}}.

Prompt templates are shown in Figure~\ref{fig:persona_generation_1}, Figure~\ref{fig:persona_generation_2}, and Figure~\ref{fig:persona_and_skill_to_problem}. Given the raw text from web page, we use Figure~\ref{fig:persona_generation_1} to prompt \llm for generating a persona. Next, we expand the persona using the prompt in Figure~\ref{fig:persona_generation_2}. Finally with the expanded persona hub, we prompt models to generate a math problem specific to each persona and skill.

\begin{figure*}[p]
\centering
\begin{tcolorbox}[
    enhanced jigsaw,
    breakable,                
    title=Text-to-persona template,  
    colback=gray!5,          
    colframe=gray!80!black,  
    fonttitle=\bfseries,
    boxrule=0.5pt,
    arc=3pt                   
]
\setlength{\parskip}{6pt}   
\setlength{\parindent}{0pt} 
  Who is likely to read the text?

  \{text\}

  Note:
  1. Your response should always start with ``Persona:''.
  2. The persona should be realistic and detailed, but don't include specific name of person.
  3. Don't include any preamble or disclaimer, but only provide the persona.
  4. Persona should be at most two sentences.
  
  Persona:
\end{tcolorbox}
\caption{Prompt template for text-to-persona generation.}
\label{fig:persona_generation_1}
\end{figure*}

\begin{figure*}[p]
\centering
\begin{tcolorbox}[
    enhanced jigsaw,
    breakable,                
    title=Persona-to-persona template,  
    colback=gray!5,          
    colframe=gray!80!black,  
    fonttitle=\bfseries,
    boxrule=0.5pt,
    arc=3pt                   
]

\setlength{\parskip}{6pt}   
\setlength{\parindent}{0pt} 
  Who is in close relationship with the given persona?

  \{persona\}

  Note:
  1. Your response should always start with "Related Persona:".
  2. The persona should be realistic and detailed, but don't include specific name of person.
  3. Don't include any preamble or disclaimer, but only provide the persona.
  4. Persona should be at most two sentences.
  
  Related Persona:
\end{tcolorbox}
\caption{Prompt template for persona-to-persona generation.}
\label{fig:persona_generation_2}
\end{figure*}

\begin{figure*}[p]
\centering
\begin{tcolorbox}[
    enhanced jigsaw,
    breakable,                
    title=Persona and skill to math problem template,  
    colback=gray!5,          
    colframe=gray!80!black,  
    fonttitle=\bfseries,
    boxrule=0.5pt,
    arc=3pt                   
]

\setlength{\parskip}{6pt}   
\setlength{\parindent}{0pt} 
  Create a math problem related to the following persona and require understanding of the following skills:

  Skills:
  \{skills\}

  Persona:
  \{persona\}

  Note:
  1. The math problem should be challenging and involve given advanced mathematical skills. Only top talents can solve it correctly.
  2. You should make full use of the persona description to create the math problem to ensure that the math problem is unique and specific to the persona.
  3. Your response should always start with "Math problem:". Your response should not include a solution to the created math problem.
  4. Your created math problem should include no more than 2 sub-problems.

  Math problem:
\end{tcolorbox}
\caption{Prompt template for persona and skill to problem generation.}
\label{fig:persona_and_skill_to_problem}
\end{figure*}




\section{Breakdown of Performance}
\label{sec:breakdown_perf}

We further provide a breakdown of the results showing impact of data formats and filtering in \rl training. \autoref{tab:q_template_det} and \autoref{tab:a_template_det} shows the impact of the question and answer formats across all tasks. In \autoref{tab:data_filter_det}, we further extend the results for models trained on unfiltered and filtered datasets.

\begin{table*}[htbp]
\centering
\resizebox{\textwidth}{!}{%
\begin{tabular}{lcccccccc}
\toprule
\textbf{Question Type} & \textbf{\mmlu} & \textbf{\mmlupro} & \textbf{\gpqad} & \textbf{\agieval} & \textbf{\supergpqa} & \textbf{\mathhard} & \textbf{\amc} & \textbf{Avg} \\
\midrule
\mcq+\open & 73.18 & \textbf{54.81} & 38.07 & \textbf{59.99} & 26.54 & \textbf{77.00} & 60.00 & 55.66\\
\open & \textbf{74.61}	&54.36	&\textbf{39.09}&	59.30	&\textbf{29.16} & 76.60	&\textbf{65.00}	&\textbf{56.87}\\
\bottomrule
\end{tabular}
}
\caption{\textbf{Impact of Question Format.} Converting all questions to open-ended format improves accuracy across benchmarks, reducing reliance on option guessing and encouraging deeper reasoning.}
\label{tab:q_template_det}
\end{table*}

\begin{table*}[htbp]
\centering
\resizebox{\textwidth}{!}{%
\begin{tabular}{lcccccccc}
\toprule
\textbf{Answer Type} & \textbf{\mmlu} & \textbf{\mmlupro} & \textbf{\gpqad} & \textbf{\agieval} & \textbf{\supergpqa} & \textbf{\mathhard} & \textbf{\amc} &  \textbf{Avg} \\
\midrule
Long & 72.77	& 52.06	& 37.06	& 56.56	& 27.44	& 72.20	&\textbf{55.00}	&53.30\\
Short & \textbf{74.22}&	\textbf{54.56}	&\textbf{39.59}	&\textbf{58.01}	&\textbf{28.39}	&\textbf{74.20}&	52.50	&\textbf{54.50}\\
\bottomrule
\end{tabular}
}
\caption{\textbf{Impact of Answer Format.} Using short-form answers improves accuracy by reducing output ambiguity and avoiding penalization from rigid reward functions in rule-based training.}

\label{tab:a_template_det}
\end{table*}

\begin{table*}[htbp]
\centering
\resizebox{\textwidth}{!}{%
\begin{tabular}{lccccccccc}
\toprule
Model & \textbf{Blend} & \textbf{\mmlu} & \textbf{\mmlupro} & \textbf{\gpqad} & \textbf{\agieval} & \textbf{\supergpqa} & \textbf{\mathhard} & \textbf{\amc} & \textbf{Avg} \\
\midrule
Qwen-2.5-32B & & 83.30	&55.10	&40.40	&62.77	&33.16	&60.55 &45.00 &54.33\\\midrule
\multirow{2}{*}{\ours-32B} &  $\mathcal{B}_{gpr\uparrow}$ & 83.57&68.83	&46.70	&73.90&	37.99 & 82.40		&67.50& 65.84\\
&  $\mathcal{B}_{f(gpr)\uparrow}$ & \textbf{83.60}	&\textbf{69.43}	&\textbf{49.75}	&\textbf{75.82}	& \textbf{38.34} &\textbf{84.00}	&\textbf{75.00}&	\textbf{67.99}\\
\bottomrule
\end{tabular}
}
\caption{\textbf{Difficulty-Based Filtering.} Filtering $\mathcal{B}_{gpr\uparrow}$ to retain only hard examples (\smash{$\mathcal{B}_{f(gpr)\uparrow}$}) yields consistent gains across all tasks, highlighting the effectiveness of selective training on challenging data.}

\label{tab:data_filter_det}
\end{table*}

\section{Data Proportion across Blends}
\label{sec:data-proportion}

\ours has been trained using the datasets shown in \autoref{tab:data_dist}. To better understand the data composition used in our reinforcement learning experiments, we report the proportion of each dataset in the six blending strategies in \autoref{tab:blend-description}, introduced in Section~\ref{sec:method}. These proportions reflect how data is distributed across different sources depending on the specific blending paradigm: data source, question type, and data usefulness.


\begin{table*}[ht]
\centering
\resizebox{\textwidth}{!}{%
\begin{tabular}{cccccccc}
\hline
\textbf{Category} & \textbf{Blend Name} & \textbf{Symbol} & \textbf{Blend Description} \\
\hline
\multirow{4}{*}{Data Source} & \multirow{2}{*}{Natural Distribution} & \multirow{2}{*}{$\mathcal{B}_{nd}$}& Ratio of number of samples in a dataset divided \\
& & & by the total number of samples in all the datasets. \\
& More Math & $\mathcal{B}_{mr\uparrow}$ & 2:1 ratio of $\mathcal{D}_{mr}$ and $\mathcal{D}_{gpr}$ \\
& More General Purpose Reasoning & $\mathcal{B}_{gpr\uparrow}$ & 2:1 ratio of $\mathcal{D}_{gpr}$ and $\mathcal{D}_{mr}$ \\
\hline
\multirow{2}{*}{Question Types} & More \mcq & $\mathcal{B}_{mcq\uparrow}$& 2:1 ratio of $\mathcal{D}_{mcq}$ and $\mathcal{D}_{open}$\\
 & More Open-Ended & $\mathcal{B}_{open\uparrow}$ & 2:1 ratio of $\mathcal{D}_{open}$ and $\mathcal{D}_{mcq}$ \\
\hline
\multirow{2}{*}{Data Usefulness}  & \multirow{2}{*}{Avg. Score} & \multirow{2}{*}{$\mathcal{B}_{score}$} & Provide weight to each source based \\
& & & on their average benchmark performances \\
\hline
\end{tabular}
}
\caption{\textbf{Overview of Data Blending Strategies.} Blends are categorized by data source, question type, and usefulness—each constructed to assess the impact of domain diversity, format variation, and task relevance on RL-based reasoning.}
\label{tab:blend-description}
\end{table*}

\begin{table*}[ht]
\centering
\resizebox{\textwidth}{!}{%
\begin{tabular}{lccccccccc}
\toprule
\textbf{Data Name} & \textbf{Type} & \textbf{$\mathcal{B}_{nd}$} & \textbf{$\mathcal{B}_{mr\uparrow}$} & \textbf{$\mathcal{B}_{mcq\uparrow}$} & \textbf{$\mathcal{B}_{open\uparrow}$} & \textbf{$\mathcal{B}_{gpr\uparrow}$} & \textbf{$\mathcal{B}_{score}$} & \textbf{$\mathcal{B}_{only\_math}$} & $\mathcal{B}_{only\_gpr}$ \\
\midrule
\mmlu         & MCQ           & 0.1696	& 0.0864	& 0.2251&	0.1159&	0.1678&	0.1296&		& 0.2542 \\
\ours-\qa & MCQ           & 0.3277 & 0.1670& 0.4349 &	0.2241 &	0.3242& 0.1731& &0.4912 \\
\textsc{Natural Reasoning}  & \open    &  0.1699	& 0.0866	& 0.1149	& 0.2231	& 0.1680	& 0.1683	&	    & 0.2546\\
NuminaMath         & \open    & 0.1484	& 0.2943	& 0.1004	& 0.1949	& 0.1516	& 0.2020	&0.4460	&  \\
\ours-\mathall    & \open   & 0.1699	& 0.3370	& 0.1149	& 0.2231	& 0.1736	& 0.1579	&0.5105	&  \\
\mathall               & \open    & 0.0145	& 0.0287	& 0.0098	& 0.0190	& 0.0148	& 0.1691	&0.0435	&  \\
\bottomrule
\end{tabular}
}
\caption{Proportion of each dataset in different blends.}
\label{tab:blend_composition}
\end{table*}

\begin{table}[h!]
\centering
\resizebox{\linewidth}{!}{%
\begin{tabular}{lcccc}
\hline
\textbf{Task Type} & \textbf{Model} & \textbf{Min} & \textbf{Max} & \textbf{Mean} \\
\midrule
\multirow{3}{*}{GPR} 
    & $\mathcal{B}_{gpr\uparrow}$   & 83.20  & 2697.80  & 385.41 \\
    & $\mathcal{B}_{only\_mr}$  & 159.60 & 9594.00  & 638.57 \\
    & \orz & 223.00 & 8221.80 &	1114.60 \\
\hline
\multirow{3}{*}{Math} 
    & $\mathcal{B}_{gpr\uparrow}$  & 170.25 & 10130.00 & 622.00 \\
    & $\mathcal{B}_{only\_mr}$   & 201.75 & 11330.25 & 730.68 \\
    & \orz & 292.00 & 12917.00 & 1257.00\\
\hline
\end{tabular}%
}
\caption{Token length statistics (Min, Max, Mean) for correct responses across task types.}
\label{tab:token_stats}
\end{table}

\section{Token Efficiency Analysis}
\label{sec:token-efficiency}

\paragraph{Token Efficiency in Correct Responses.} Understanding not only whether a model answers correctly but also how efficiently it reasons is critical in real-world deployments, especially for reducing inference cost and latency. To this end, we analyze the token lengths of correct responses generated by models trained under different data blending strategies.

\autoref{tab:token_stats} presents the minimum, maximum, and mean number of tokens used in correct answers across two task types: General Purpose Reasoning (GPR) and Math. We compare three models: (1) $\mathcal{B}_{gpr\uparrow}$ (multi-domain training), (2) $\mathcal{B}_{only\_math}$ (math-only training), and (3) \orz (a strong math-centric baseline model).

Across \gpr tasks, $\mathcal{B}_{gpr\uparrow}$ produces the most concise correct responses, with a mean of 385 tokens—39.6\% fewer than $\mathcal{B}_{only\_mr}$ and 65.4\% fewer than \orz. This suggests that training with multi-domain corpora equips the model to reason more efficiently in less structured tasks, avoiding unnecessarily verbose responses.

On math benchmarks, where detailed step-by-step derivations are essential, all models naturally generate longer outputs. However, $\mathcal{B}_{gpr\uparrow}$ still demonstrates adaptability, producing appropriately longer responses compared to GPR, while keeping the output concise relative to $\mathcal{B}_{only\_math}$ and \orz. This behavior underscores the ability of multi-domain trained models to dynamically adjust their reasoning strategy and verbosity based on task requirements.

Interestingly, \orz exhibits the longest response lengths across both GPR and math tasks. While this aligns with its design as a reasoning-heavy model, it also reflects less efficiency—potentially generating unnecessarily long chains of thought, particularly in domains outside its training focus.

In summary, the token efficiency analysis reveals that $\mathcal{B}_{gpr\uparrow}$ achieves a favorable trade-off between accuracy and brevity, tailoring its reasoning depth to the complexity of the task. This reinforces the value of diverse, multi-domain training in promoting adaptable and cost-efficient language models.



\begin{figure}
  \centering
  \includegraphics[width=\linewidth]{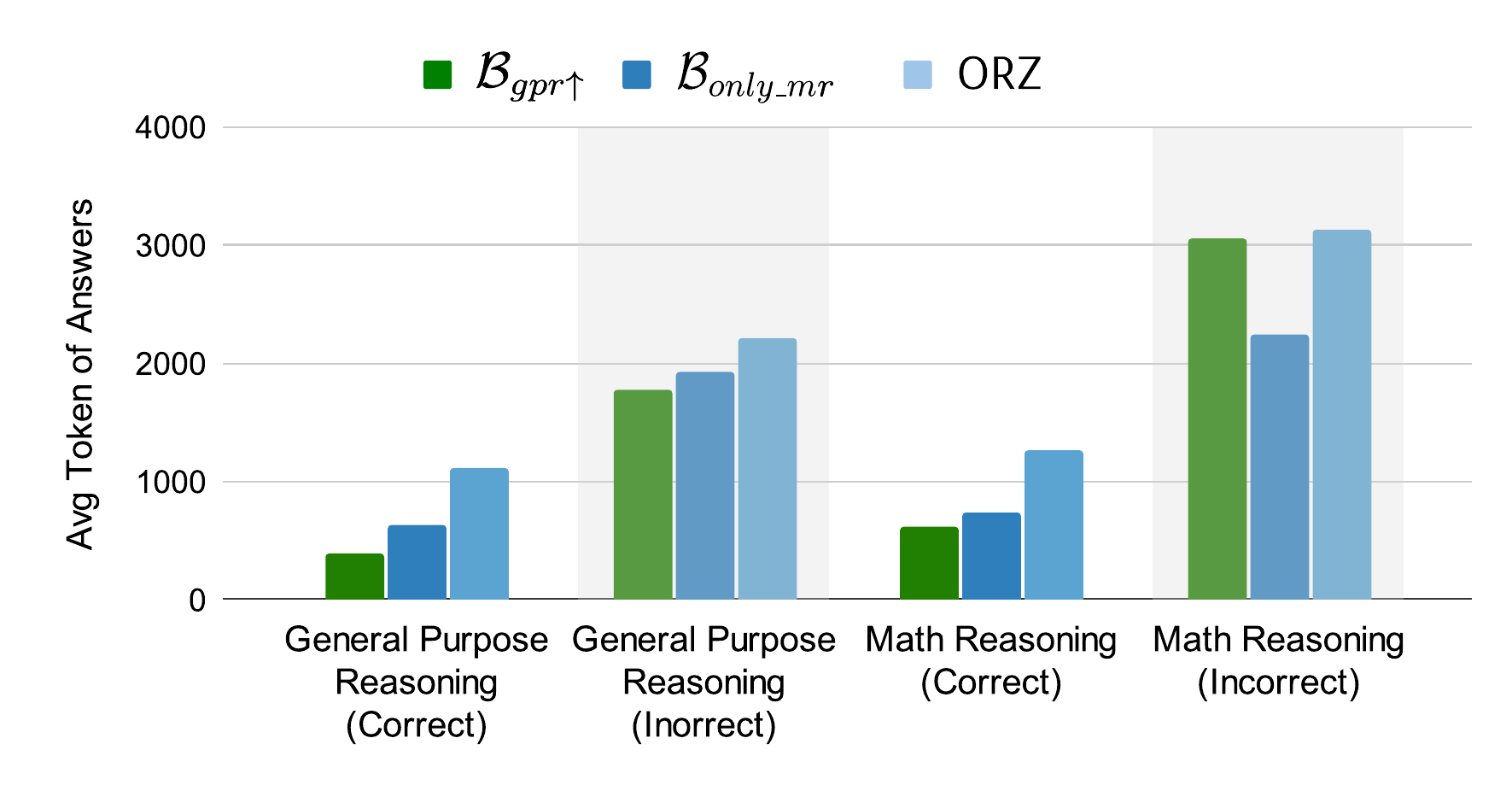}
  \caption{Average token lengths of correct and incorrect responses across general-purpose and math reasoning tasks for models trained on $\mathcal{B}_{gpr\uparrow}$, $\mathcal{B}_{only\_mr}$, and \orz.}
  \label{fig:incorr_tokens}
\end{figure}
\paragraph{Thinking Long vs Thinking Accurate. } Recent studies such as DeepScaler \citep{deepscaler2025} have noted that incorrect answers often exhibit longer trajectories, leading to wasted computation and less efficient learning. Echoing this observation, we analyze the average token lengths of correct and incorrect responses for models trained on different blends: $\mathcal{B}_{gpr\uparrow}$, $\mathcal{B}_{only\_mr}$, and \orz.

As shown in \autoref{fig:incorr_tokens}, incorrect responses are consistently and substantially longer than correct ones—by 3.6$\times$ on average. This pattern holds across both general-purpose and math reasoning tasks, suggesting that verbose reasoning does not guarantee correctness. In fact, longer responses often reflect the model's uncertainty, overthinking, or repetitive CoT traces, rather than productive deduction.

\section{Sub-category Accuracy Analysis}
\label{sec:subcat_abl}

\begin{figure*}
  \centering
  \includegraphics[width=\textwidth]{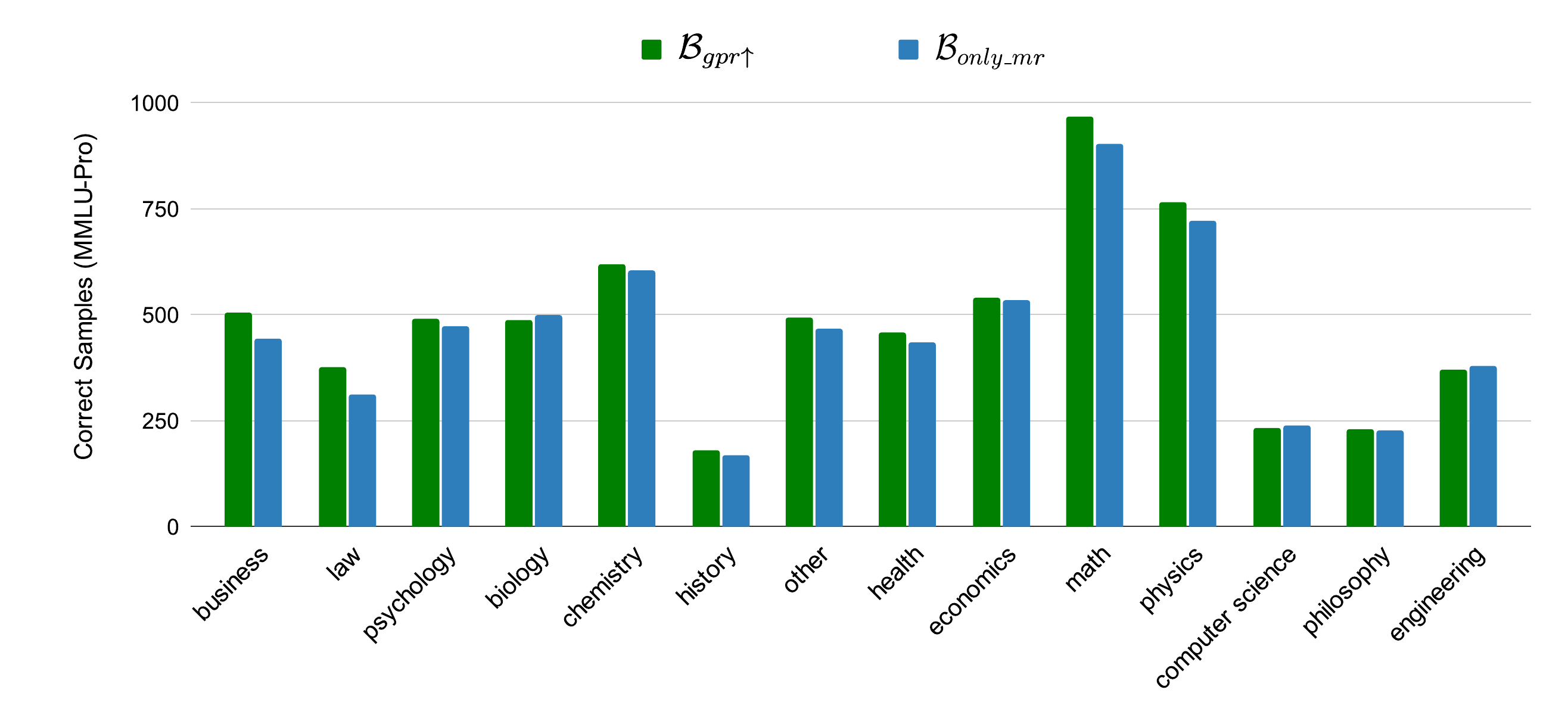}
  \caption{\textbf{Sub-category Accuracy Comparison across \mmlupro Domains.} The $\mathcal{B}_{gpr\uparrow}$ blend consistently outperforms $\mathcal{B}_{only\_mr}$ in a wide range of non-math reasoning categories such as business, law, psychology, and economics. Surprisingly, it also slightly surpasses the math-specialized blend in the \mmlupro math category, highlighting the generalizability and versatility of multi-domain training.}
  \label{fig:mmlu_abl}
  \vspace{-3mm}
\end{figure*}

\begin{figure}
  \centering
  \includegraphics[width=\linewidth]{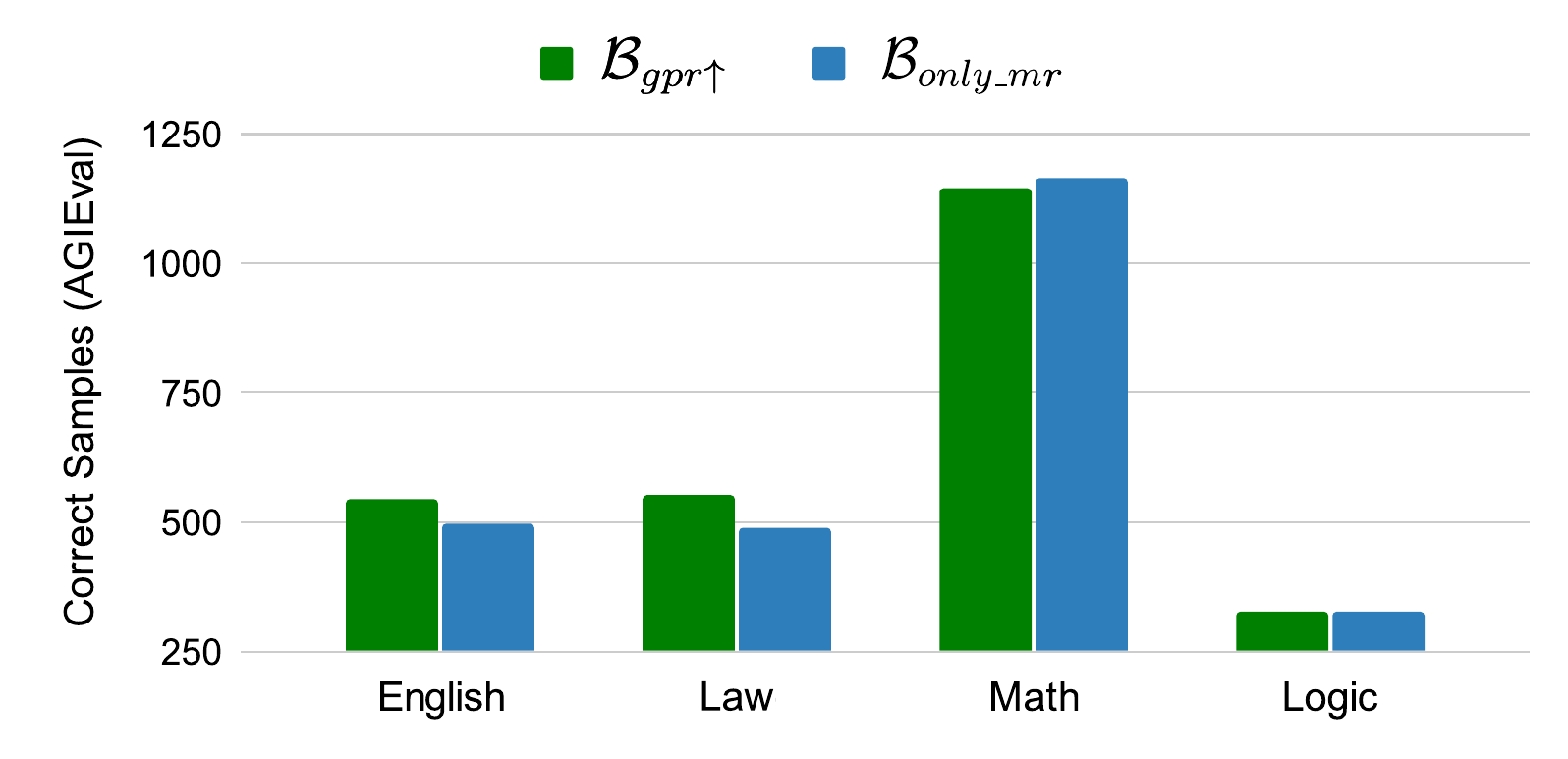}
  \caption{\textbf{Sub-category Accuracy Comparison across \agieval.} While $\mathcal{B}_{only\_mr}$ performs marginally better in the math, $\mathcal{B}_{gpr\uparrow}$ achieves stronger results in non-math domains.}
  \label{fig:agieval_abl}
  \vspace{-2mm}
\end{figure}

To further support our observation that multi-domain training improves general-purpose reasoning while remaining competitive on math tasks, we analyze the number of correct responses across sub-categories in \mmlupro and \agieval. \autoref{fig:mmlu_abl} and \autoref{fig:agieval_abl} show the count of correct answers produced by $\mathcal{B}_{gpr\uparrow}$ and $\mathcal{B}_{only\_math}$ across their respective sub-domains.

On \mmlupro, $\mathcal{B}_{gpr\uparrow}$ consistently outperforms $\mathcal{B}_{only\_math}$ across non-math reasoning categories such as business, law, psychology, chemistry, and economics. Notably, it achieves relative improvements of +20.58\% in law and +13.26\% in business. Surprisingly, $\mathcal{B}_{gpr\uparrow}$ also performs better in the math category (+7.2\%), despite not being trained exclusively on mathematical data. This may be attributed to the nature of {\mmlupro}’s math problems, which are college-level and benefit from a combination of symbolic and heuristic reasoning—skills reinforced through exposure to diverse domains.

In contrast, the \agieval benchmark (shown in \autoref{fig:agieval_abl}) features Olympiad-level math questions that are more abstract and complex. Here, $\mathcal{B}_{only\_math}$ has a slight edge (+1.8\%) in the math category, which aligns with its domain-specific training. However, $\mathcal{B}{gpr\uparrow}$ demonstrates stronger performance in symbolic and language-heavy domains, showing +13.06\% improvement in Law and +9.88\% in English. Averaged across all non-math reasoning categories, $\mathcal{B}_{gpr\uparrow}$ achieves a +8.6\% relative gain over $\mathcal{B}_{only\_math}$, reinforcing its advantage in general-purpose and real-world reasoning tasks.

\begin{figure}
  \centering
  \includegraphics[width=\linewidth]{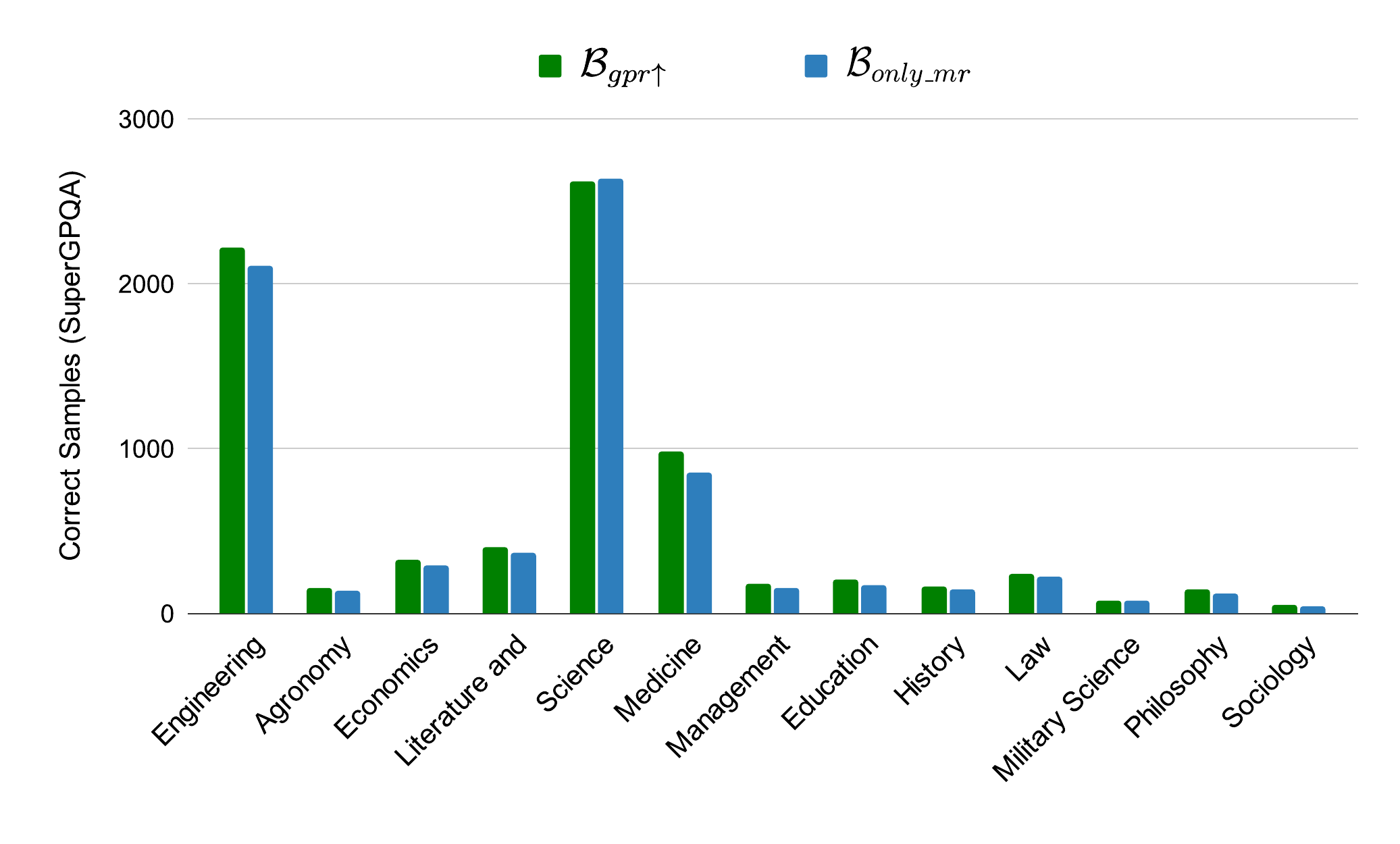}
  \caption{\textbf{Sub-category Accuracy Comparison across \supergpqa.} The $\mathcal{B}_{gpr\uparrow}$ blend consistently outperforms $\mathcal{B}_{only\_mr}$ in a wide range of non-math reasoning categories except the science category which consists of fields like mathematics, physics, astronomy, chemistry etc.---highlighting the generalizability and versatility of multi-domain training.}
  \label{fig:supergpqa_abl}
\end{figure}

A similar trend is observed in the \supergpqa sub-category analysis shown in \autoref{fig:supergpqa_abl}. $\mathcal{B}_{gpr\uparrow}$ significantly outperforms $\mathcal{B}_{only\_math}$ across nearly all categories—especially in engineering, agronomy, economics, education, law, and philosophy. The only exception is the ``Science'' category, which includes math-heavy disciplines like physics, chemistry, and astronomy, where both blends perform comparably. This further highlights that multi-domain training with $\mathcal{B}_{gpr\uparrow}$ enhances reasoning across a broad spectrum of fields, achieving strong generalization even in real-world, professional domains that fall outside traditional math tasks.

\section{Generalizability Across Model Architectures} \label{sec:general_model}
To verify that the efficacy of \ours is not limited to a single model architecture, we extended our evaluation to the Nemotron-H \citep{nvidia2025nemotronhfamilyaccurateefficient} model family. We conducted a controlled experiments only for 200 steps using our two best performing data blends: $\mathcal{B}_{only\_mr}$, a domain-specific ``Math-Only'' blend and $\mathcal{B}_{gpr\uparrow}$, the ``CrossThink'' mixed-domain blend (utilizing a 2:1 ratio of general-purpose to math domains).

\begin{table*}[t]
\centering
\resizebox{\textwidth}{!}{%
\begin{tabular}{lcccccccc}
\toprule
\textbf{Model Blend} & \textbf{\mmlu} & \textbf{\mmlupro} & \textbf{\gpqad} & \textbf{\agieval} & \textbf{\supergpqa} & \textbf{\mathhard} & \textbf{\amc} & \textbf{Avg} \\ \midrule
$\mathcal{B}_{only\_mr}$ & 68.48 & 48.58 & 34.01 & 57.94 & 23.10 & \textbf{75.36} & \textbf{51.50} & 46.42 \\
$\mathcal{B}_{gpr\uparrow}$ & \textbf{69.94} & \textbf{49.66} & \textbf{39.09} & \textbf{58.23} & \textbf{24.60} & 70.88 & 48.17 & \textbf{48.30} \\ \bottomrule
\end{tabular}%
}
\caption{\textbf{Expanding to $\mathcal{M}$=Nemotron-H:} The relative trend confirms that \ours improves general reasoning (MMLU, GPQA, etc.) compared to the Math-Only baseline.}
\label{tab:nemotron_perf}
\end{table*}

As observed in \autoref{tab:nemotron_perf}, the CrossThink blend demonstrates a strong capacity for transfer learning. While the math-specific scores (\textsc{math-500}, \textsc{amc23}) are slightly lower than the Math-Only baseline—an expected outcome given that the CrossThink blend contains fewer math samples due to the 2:1 mixing ratio—the model significantly gains in general reasoning capabilities. Specifically, CrossThink outperforms the baseline on general-purpose reasoning tasks, achieving a 1.88\% improvement in the Reasoning Average. This suggests that mixing domains helps the model generalize better to unseen reasoning tasks, even within a new architecture.

Beyond performance metrics, we analyzed the inference efficiency of the approach on the Nemotron-H architecture. We calculated the average number of tokens generated for correct responses across seven benchmarks.

\begin{table}[t]
\centering
\resizebox{\linewidth}{!}{%
\begin{tabular}{lccc}
\toprule
\textbf{Task} & \textbf{$\mathcal{B}_{only\_mr}$} & \textbf{$\mathcal{B}_{\uparrow gpr}$} & \textbf{Reduction} \\ \midrule
\mmlu & 387.69 & 313.39 & 19.2\% \\
\agieval & 525.10 & 432.25 & 17.7\% \\
\mmlupro & 772.27 & 495.56 & 35.8\% \\
\gpqad & 1124.62 & 656.15 & 41.7\% \\
\supergpqa & 421.52 & 306.13 & 27.4\% \\
\mathhard & 551.96 & 460.72 & 16.5\% \\
\amc & 557.42 & 438.66 & 21.3\% \\ \midrule
\textbf{Average} & \textbf{620.08} & \textbf{443.26} & \textbf{28.5\%} \\ \bottomrule
\end{tabular}%
}
\caption{\textbf{Average token counts for correct responses (Nemotron-H).} \ours significantly reduces the verbosity of the model across all benchmarks, achieving a 28.5\% average reduction in token generation for correct answers.}
\label{tab:nemotron_tokens}
\end{table}

As shown in \autoref{tab:nemotron_tokens}, the \ours model generates 28.51\% fewer tokens on average compared to the baseline while maintaining or improving accuracy on general reasoning tasks. This confirms that the efficiency gains observed in Qwen-2-7B are architectural-agnostic: \ours consistently encourages more concise reasoning paths without sacrificing solution correctness.

\section{Additional Discussion}

\paragraph{Why \ours is applied directly on the base model? } In this work, we apply RL directly to pretrained language models rather than to fine-tuned models. Although RL on SFT models is often considered more stable due to instruction-following priors and task structure acquired during supervised training \citep{deepseekai2025deepseekr1incentivizingreasoningcapability}, prior findings show that this choice does not lead to consistent benefits.

Several studies report that SFT can reduce model entropy, which is important for effective exploration during reinforcement learning \citep{pmlr-v267-chu25c}, and that applying RL methods such as GRPO \citep{deepseek-math} to already aligned or instruction-tuned models can hurt performance \citep{chen2025sft}. Other work observes improvements when RL is applied after heavy SFT \citep{liu2025acereasonnemotron11advancingmath}. These mixed outcomes indicate high variance in the effectiveness of RL on SFT models, making it difficult to isolate the contribution of reinforcement learning itself.

We therefore focus on RL applied to the pretrained base model to study the emergence of reasoning abilities without strong instruction-following priors and to avoid confounding effects introduced by SFT. This choice is further motivated by recent work showing that self-learning on base models can yield substantial gains in reasoning capabilities across diverse tasks \citep{zeng2025simplerl, hu2025openreasonerzeroopensourceapproach, wang2025octothinker}.

\paragraph{Robustness across Decoding Strategies (Greedy vs. Pass@1[8]).} In addition to greedy decoding, we evaluated two of our strongest models using pass@1[8] with stochastic decoding (temperature = 0.6, top\_p = 0.95, k = 8). The results are summarized in \autoref{tab:pass1_results}.

\begin{table*}[t]
\centering
\resizebox{\linewidth}{!}{%
\begin{tabular}{lccccccccc}
\toprule
Model & \mmlu & \mmlupro & \gpqad & \agieval & \supergpqa & \mathhard & \amc & \aime & Avg \\
\midrule
$\mathcal{B}_{only\_mr}$ & 72.26& 53.40	& 34.54	& 61.62	& 22.35	& 76.88	& 62.00	& 16.68	& 49.97\\
$\mathcal{B}_{gpr\uparrow}$ & 73.82	&57.01&	37.86&	64.00	&25.14	&76.03	&61.11	&15.27	&51.28 \\
\bottomrule
\end{tabular}
}%
\caption{Pass@1[8] performance with stochastic decoding (temperature = 0.6, top\_p = 0.95, $k=8$).}
\label{tab:pass1_results}
\end{table*}

Under pass@1[8], the $\mathcal{B}_{gpr\uparrow}$ model achieves the highest overall average score, outperforming the $\mathcal{B}_{only\_mr}$ model. The gains are most pronounced on general-purpose reasoning benchmarks, including \mmlu, \mmlupro, \gpqad, etc., where $\mathcal{B}_{gpr\uparrow}$ consistently improves over $\mathcal{B}_{only\_mr}$. On math-focused tasks such as \mathhard, \amc, and \aime, the two models exhibit comparable performance, with $\mathcal{B}_{only\_mr}$ showing a small advantage on some benchmarks. These results mirror the trends observed with greedy decoding and further support that the multi-domain blend improves broad reasoning performance without sacrificing math capability.

\section{Extended Related Work}

\paragraph{Self-Learning beyond Math. } High-quality training data are crucial for scalable Reasoner-Zero training. Most of the recent works emphasize mathematical benchmark-centric data (AMC, AIME, Math, Olympiads, and AoPS) for reinforcement learning \citep{OpenReasonerZero2025, aggarwal2025l1controllinglongreasoning, trung-etal-2024-reft, ye2025limoreasoning, zeng2025simplerlzooinvestigatingtamingzero} as designing verifiable rewards is much easier for math tasks. They exclude problems such as multiple choice and proof-oriented problems which reduces the answer space diversity. \mcq type of questions are important for \mmlu and other non-reasoning centric tasks. Recently, \cite{generalreasoner, su2025crossingrewardbridgeexpanding} attempted to address this with a model-based verifier to handle diversity in the answer space. However, as discussed in previous works, \llm-as-a-Judge may suffer from pitfalls of reward hacking \citep{deepseekai2025deepseekr1incentivizingreasoningcapability, weng2024rewardhack, wen2024languagemodelslearnmislead} and further diverge the model more from the correct reasoning processes. Additionally, they do not show any analysis to estimate the contribution of each domain in the final task performance. Despite training on diverse domains, \ours offers simple, scalable and robust reward estimation without any external reward model. For a rule-based reward model, the format of input data and the final answer is crucial and largely underexplored. Furthermore, their additional sources of data synthesis approach has no details making it infeasible to scale for domains other than math. The kind of data and the ratio of each type of data important for the overall improvement of {\llm}s across multiple benchmarks have yet to be explored.

\end{document}